\documentclass{article}

\usepackage[preprint]{neurips_2026}

\usepackage[utf8]{inputenc} 
\usepackage[T1]{fontenc}    
\usepackage{hyperref}       
\hypersetup{colorlinks=true, linkcolor=red, citecolor=black, urlcolor=black, filecolor=black, pdfborder={0 0 0}}
\usepackage{url}            
\usepackage{booktabs}       
\usepackage{amsfonts}       
\usepackage{nicefrac}       
\usepackage{microtype}      
\usepackage{xcolor}         
\usepackage{amsmath}
\usepackage[most]{tcolorbox}
\usepackage{xcolor}
\usepackage{makecell}
\usepackage{subcaption}
\usepackage{multirow}
\usepackage{makecell}
\usepackage[table]{xcolor}
\usepackage{threeparttable}
\usepackage{array}

\newcommand{\finding}[2]{%
\begin{tcolorbox}[
    enhanced,
    colback=yellow!12,
    colframe=red!55!black,
    boxrule=0.4pt,
    arc=0pt,
    left=2mm,
    right=2mm,
    top=0.6mm,
    bottom=0.6mm,
    boxsep=0pt
]
{\color{red!65!black}\textit{Finding~#1.} #2}
\end{tcolorbox}
}

\definecolor{PanelGray}{HTML}{F3F4F6}
\definecolor{OursGreen}{HTML}{E7F5E7}
\definecolor{GainBlue}{HTML}{EAF2FF}
\definecolor{TopGray}{HTML}{F8F9FA}
\definecolor{MCOursGreen}{HTML}{E7F5E7}

\newcommand{\openaiicon}{\raisebox{-0.12em}{\includegraphics[height=0.85em]{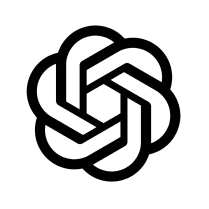}}}
\newcommand{\anthropicicon}{\raisebox{-0.12em}{\includegraphics[height=0.85em]{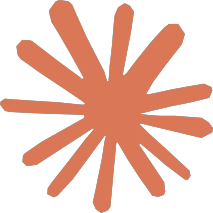}}}
\newcommand{\googleicon}{\raisebox{-0.12em}{\includegraphics[height=0.85em]{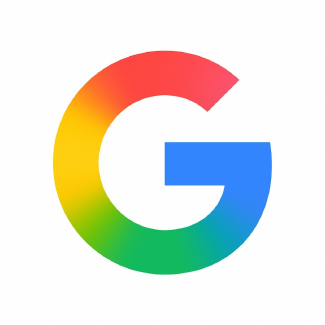}}}
\newcommand{\qwenicon}{\raisebox{-0.12em}{\includegraphics[height=0.85em]{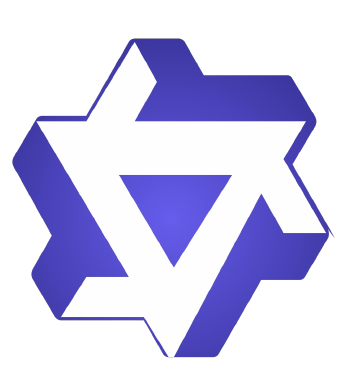}}}

\definecolor{OursGreen}{HTML}{E7F5E7}
\definecolor{GainBlue}{HTML}{EAF2FF}
\definecolor{PanelGray}{HTML}{F2F2F2}

\definecolor{MCOursGreen}{HTML}{E7F5E7}
\definecolor{MCGainBlue}{HTML}{EAF2FF}
\definecolor{MCPanelGray}{HTML}{E9ECEF}

\title{LLMs Know When They Know, but Do Not Act on It: A Metacognitive Harness for Test-time Scaling}

\author{%
  Qi Cao \\
  University of California, San Diego\\
  \texttt{q9cao@ucsd.edu} \\
  \And
  Yufan Wang \\
  University of California, San Diego\\
  \texttt{yuw313@ucsd.edu} \\
  \And
  Peijia Qin \\
  University of California, San Diego\\
  \texttt{pqin@ucsd.edu} \\
  \And
  Shuhao Zhang \\
  University of California, San Diego\\
  \texttt{shz127@ucsd.edu} \\
  \And
  Pengtao Xie \\
  University of California, San Diego\\
  \texttt{p1xie@ucsd.edu} \\
}

\begin{document}

\maketitle

\begin{abstract}
Large language models (LLMs) often expose useful signals of self-monitoring: before solving a problem, they can estimate whether they are likely to succeed, and after solving it, they can judge whether their answer is likely to be correct. However, these signals are typically measured or elicited in isolation, rather than used to control inference. In this work, we ask whether LLMs possess latent metacognitive ability that can be turned into effective test-time control. Inspired by the Nelson--Narens theory from cognitive psychology, we propose a metacognitive harness that separates monitoring from reasoning. For each problem, the model first reports a pre-solve feeling-of-knowing (FOK) signal; after each solve attempt, it reports a post-solve judgment-of-learning (JOL) signal. Rather than treating these signals as passive confidence estimates, the harness turns them into an explicit control interface for reasoning: it decides when to trust the current solution, when to retry with compact metacognitive feedback, and when to pass multiple attempts to a final aggregator. Across text, code, and multimodal reasoning benchmarks, our harness substantially improves a fixed Claude Sonnet-4.6 base model without parameter updates or benchmark-specific fine-tuning. On the evaluated public benchmark snapshots, it raises pooled accuracy from 48.3 to 56.9 and exceeds the strongest listed leaderboard entries on the three primary evaluation settings: HLE-Verified, LiveCodeBench v6, and R-Bench-V. These results suggest that strong LLMs may already possess useful metacognitive ability, but require an explicit control harness to act on it during reasoning.
\end{abstract}

\section{Introduction}

\begin{figure}[t]
    \centering
    \includegraphics[width=\linewidth]{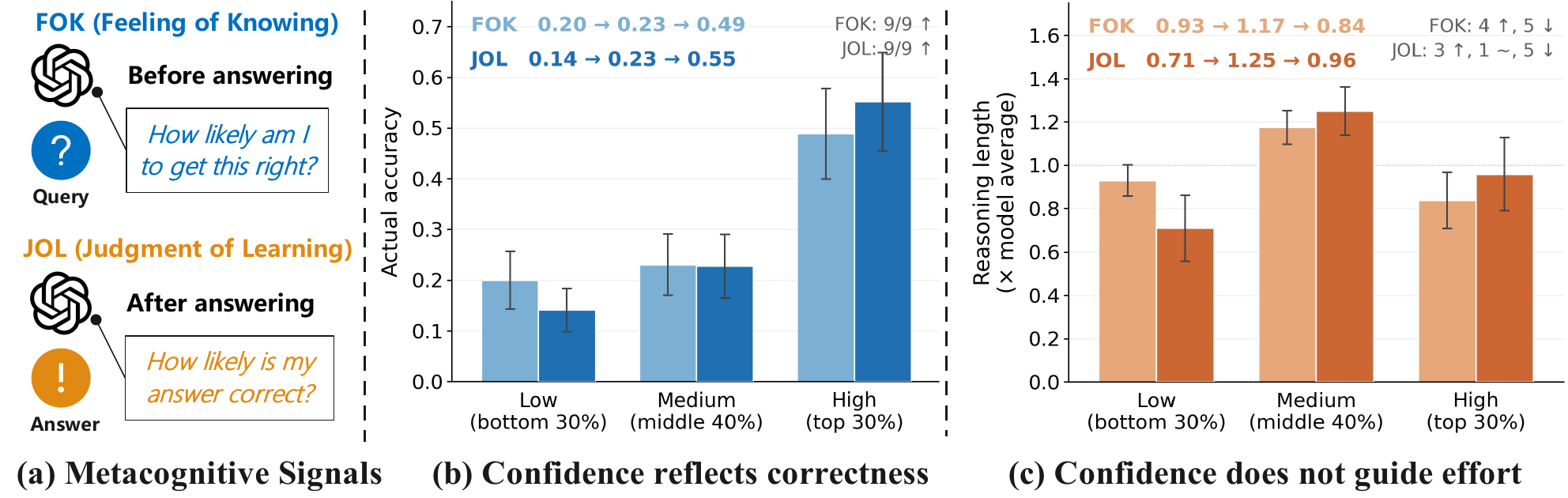}
    \caption{
\textbf{LLMs exhibit metacognitive signals, but do not use them to control reasoning.}
(a) We directly prompt each LLM to report a scalar self-assessment in $[0,1]$ before answering, denoted as FOK (\emph{Feeling of Knowing}), and after answering, denoted as JOL (\emph{Judgment of Learning}). 
(b) These self-reported scores are meaningfully correlated with actual correctness: examples with higher FOK/JOL scores achieve higher accuracy, suggesting that LLMs possess usable metacognitive monitoring signals. 
(c) However, these signals do not translate into adaptive reasoning control: higher-confidence examples do not consistently receive more or less reasoning effort, measured by reasoning length normalized by each model's average. Arrow counts summarize the per-model comparison between the high- and low-confidence bands across the nine evaluated LLMs: $\uparrow$ means High $>$ Low and $\downarrow$ means High $<$ Low.
Details for the 9 evaluated LLMs are reported in Appendix~\ref{sec:appendix-per-model}.
}
    \label{fig:harness_engineering}
\end{figure}

\begin{quote}
``Humans steer. Agents execute.''  --- \textit{OpenAI, ``Harness Engineering''}~\citep{lopopolo2026harnessengineering}
\end{quote}

Large language models (LLMs) have become increasingly capable in recent years \citep{achiam2023gpt4,wei2022chain,snell2025scaling}. They can solve a wide range of tasks, including coding \citep{roziere2023code}, mathematical reasoning \citep{lewkowycz2022solving,gao2023pal}, question answering \citep{achiam2023gpt4,yao2023react}, and agentic tasks that require planning and tool use \citep{yao2023react,schick2023toolformer,anthropic2025effectiveharnesses}. Across many challenging benchmarks, LLMs have shown strong performance, often approaching or even surpassing human performance in specific domains \citep{achiam2023gpt4,lewkowycz2022solving,snell2025scaling}. As a result, the role of humans is shifting from solving problems step by step to steering and controlling how these systems operate. Recent discussions have made this view increasingly explicit: OpenAI frames this layer as harness engineering~\citep{lopopolo2026harnessengineering}, Anthropic highlights the importance of harness engineering for agent behavior~\citep{anthropic2025effectiveharnesses}, and LangChain emphasizes harness design and human-in-the-loop control for reliable agent systems~\citep{trivedy2026agentharness,langchain2026harnesscapabilities}. Together, these developments suggest that, as LLMs become stronger, an important human role is to design the right harnesses that help models use their abilities more effectively.

One such idea, which remains underexplored, is metacognition. A growing body of work suggests that LLMs expose useful signals about their own uncertainty and correctness~\citep{kadavath2022language,xiong2024can,didolkar2024metacognitive,yuan2024fact}. These studies establish an important premise: self-reported confidence and related metacognitive signals can correlate with actual success. However, they mostly treat metacognition as a diagnostic object---something to elicit, measure, calibrate, or analyze---rather than as a mechanism for controlling reasoning. Figure~\ref{fig:harness_engineering} makes this gap explicit: LLMs can report confidence scores before solving and after answering; these scores predict empirical correctness, but the model's own reasoning effort does not reliably adapt to them. In other words, knowing that it is uncertain does not by itself make the model retry, verify, or allocate more computation. This raises our central question: can metacognitive signals be turned into a control interface that regulates reasoning itself?

We answer this question by turning metacognition from a diagnostic signal into a control interface. Inspired by the Nelson--Narens theory from cognitive psychology~\citep{nelson1990metamemory,flavell1979metacognition}, we treat the model's self-monitoring signals not as endpoints to be measured, but as inputs to an inference-time controller. A lightweight diagnosis stage first tests, on a small disjoint anchor set, whether these signals are discriminative and calibrated enough to support control. It then fits a model-specific decision rule that determines when a current attempt should be trusted and when additional computation should be allocated. At test time, the resulting harness performs selective test-time scaling: reliable attempts stop early, while uncertain but promising cases receive additional reasoning attempts and final aggregation. Across STEM, coding, and multimodal reasoning benchmarks, this simple control layer substantially improves a fixed Claude Sonnet-4.6 base model without parameter updates or benchmark-specific fine-tuning, achieving state-of-the-art performance on all evaluated public leaderboards.

Our contributions are summarized as follows:
\begin{itemize}
    \item We propose a metacognitive harness that turns self-monitoring signals into inference-time control. Building on the metacognition theory of Nelson and Narens~\citep{nelson1990metamemory,flavell1979metacognition}, the harness uses the model's own metacognitive feedback to decide when to trust, retry, stop, and aggregate reasoning attempts.

    \item We introduce a lightweight metacognitive diagnosis for measuring whether an LLM exposes signals that are usable for control. The diagnosis evaluates discrimination and calibration on a small disjoint anchor set, and fits a model-specific decision rule that converts metacognitive feedback into retry and stopping decisions.

    \item We demonstrate that metacognitive control improves reasoning without changing the base model. With a fixed Claude Sonnet-4.6 model, our harness achieves state-of-the-art performance on all evaluated public leaderboards across STEM, code, and multimodal reasoning. Ablations show that the gains depend on calibrated decision rules, directed retry, and context management, rather than uniform sampling or longer single-trajectory reasoning.
\end{itemize}

\section{Related Work}
\paragraph{Metacognition in psychology.}
In cognitive psychology, metacognition broadly refers to the monitoring and regulation of one’s own cognitive processes rather than cognition at the object level alone \citep{flavell1979metacognition}. A particularly influential account is the monitoring-and-control framework of Nelson and Narens \citep{nelson1990metamemory}, in which meta-level assessments inform object-level decisions such as whether to persist, revise, terminate, or allocate additional effort. Two judgments are especially relevant here. \emph{Feeling of knowing} (FOK) is a prospective estimate of whether one is likely to know or retrieve the correct answer \citep{hart1965memory,reder1992what}, while \emph{judgment of learning} (JOL) is a more retrospective estimate of whether a produced response is likely to be correct or well learned \citep{nelson1990metamemory,metcalfe1994metacognition}. This distinction is central to our formulation, where pre-attempt and post-attempt judgments are treated as control signals.

\paragraph{Metacognition in LLMs.}
Prior work shows that LLMs can express useful uncertainty and self-evaluative signals \citep{kadavath2022language,xiong2024can,yang2024verbalized}. More recent studies investigate explicit metacognition through skill awareness \citep{didolkar2024metacognitive}, self-aware intervention \citep{tan2025tuningfree}, decoupled metacognitive evaluation \citep{wang2025decoupling}, and intrinsic metacognitive signals in internal states \citep{ma2025intrinsic}. These works suggest that LLMs may already possess meaningful metacognitive ability, but they mainly focus on eliciting, calibrating, or measuring such signals rather than using them to regulate reasoning behavior during inference.

\paragraph{Self-refine, verifiers, and test-time control.}
Another related line studies whether test-time control can improve outputs without parameter updates. Self-Refine iteratively critiques and revises the current answer~\citep{madaan2023selfrefine}, while Reflexion stores verbal reflections from task feedback for later trials~\citep{shinn2023reflexion}. Verifier- and reward-model-based methods instead improve reasoning by scoring candidate answers or intermediate steps, for example through outcome verifiers, process reward models, or learned rerankers~\citep{cobbe2021training,uesato2022solving,lightman2023lets}. These methods are powerful for selecting among generated candidates, but they typically rely on an external scoring model or a fixed candidate pool. A separate line improves reasoning by allocating more inference-time compute, for example through chain-of-thought prompting, self-consistency, and adaptive test-time scaling~\citep{wei2022chain,wang2023self,snell2025scaling}. Our work differs in that it uses the model's own prospective and retrospective metacognitive signals to control generation itself: deciding when to retry, when to stop, and how to manage context across attempts.

\section{Method}

We propose a \emph{metacognitive harness} for selective test-time scaling. The key idea is simple: rather than allocating the same amount of reasoning compute to every question, we use the model's own metacognitive signals to decide whether additional inference is likely to be useful. Our framework has two stages. First, we perform \emph{metacognition diagnosis}, where we measure whether a model produces reliable self-assessment signals and fit a retry rule from these signals. Second, we apply this fitted rule at test time to selectively allocate additional reasoning budget only to questions that appear worth retrying.

\subsection{Metacognitive Signals}

For each question, the model is prompted to produce two scalar self-assessment signals in $[0,1]$.

\paragraph{Feeling of Knowing (FOK).}
Before attempting any reasoning, the model is asked to provide a \emph{feeling of knowing} score, which reflects its initial belief about whether it knows how to solve the problem. This stage is intended to capture pre-solution intuition rather than partial derivation, so the prompt explicitly forbids computation, step-by-step reasoning, or partial answers.

\paragraph{Judgment of Learning (JOL).}
After producing a solution, the model outputs a second score, which we refer to as a \emph{judgment of learning} (JOL). Operationally, this signal acts as a post-answer confidence estimate: it reflects how certain the model is that its produced answer is correct.

Thus, for a question $x$ (and an image $v$ when applicable), a single attempt produces
\[
(\mathrm{FOK}, c, a, \mathrm{JOL}),
\]
where $a$ is the answer and $c$ is the reasoning trace. FOK is computed once before solving, while JOL is emitted after each attempt.

\subsection{Metacognition Diagnosis}
\label{sec:metacog-diag}

Before using metacognitive signals for test-time scaling, we perform a
lightweight diagnosis for each candidate model on a small \emph{anchor set}
of 100 reasoning problems spanning text, code, and
multimodal domains. The anchor set is disjoint from all downstream
benchmarks, so the diagnosis does not contaminate the final evaluation.

\paragraph{Anchor-based reliability diagnosis.}
For each model, we collect triples
$(\mathrm{FOK}_i,\mathrm{JOL}_i,y_i)$, where $y_i$ indicates whether the single-pass answer is
correct. We use these triples to assess whether the raw metacognitive signals are reliable control variables, measuring both their ability to separate correct from incorrect answers and their calibration to empirical accuracy.

\paragraph{Metacognitive decision function.}
Although $\mathrm{FOK}$ and $\mathrm{JOL}$ are model-agnostic self-reports,
different models may use these signals with different semantics, scales,
and failure modes. We therefore do not impose a hand-crafted rule, such as
a fixed $\mathrm{JOL}$ threshold or a universal combination of
$\mathrm{FOK}$ and $\mathrm{JOL}$. Instead, for each model $m$, we use the
anchor triples to learn a lightweight metacognitive decision function
\[
    g_m(\mathrm{FOK},\mathrm{JOL}) \rightarrow \widehat{y}_m \in \{0,1\},
\]
implemented as a support vector machine (SVM). This function defines a
model-specific decision boundary over pre-solve and post-solve
self-assessments, specifying when model $m$ should trust its current
answer and when the downstream control policy should allocate additional
test-time computation. Thus, the input representation remains black-box
and model-agnostic, while the control rule is adapted to the
metacognitive behavior of each individual model. The diagnostic metrics,
SVM training details, and reliability rubric are provided in
Appendix~\ref{app:metacog-diag-extended}; the resulting per-model
diagnoses are reported in Section~\ref{sec:finding-metacog}.

\subsection{Metacognitive Harness}
\label{sec:metacog-harness}

\begin{figure}[t]
    \centering
    \includegraphics[width=\linewidth]{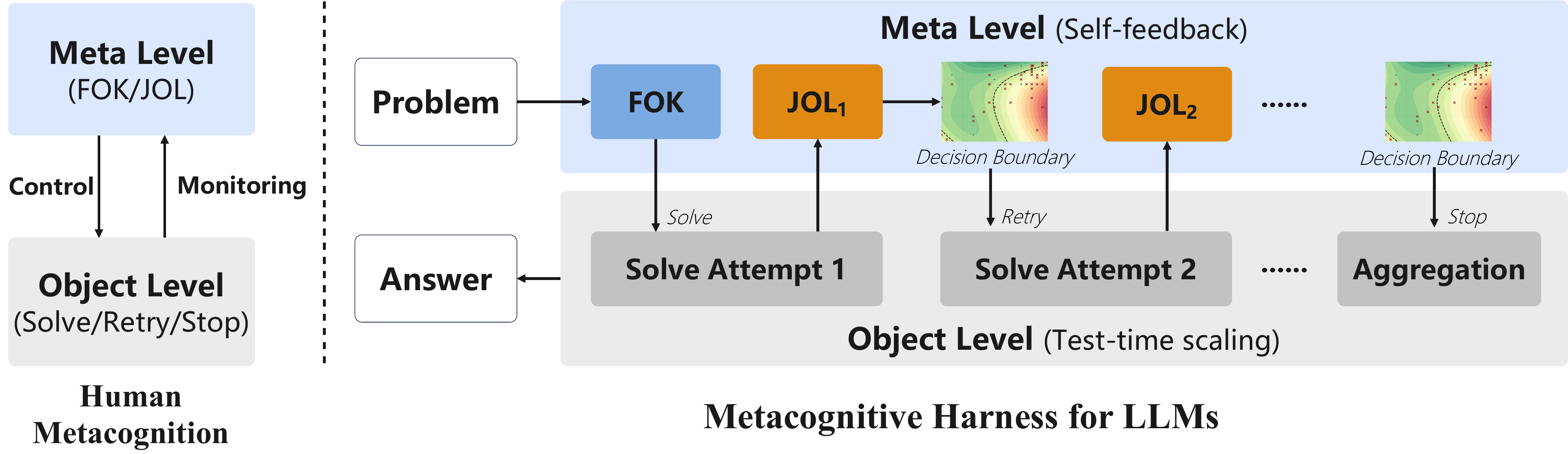}
    \caption{
\textbf{Metacognitive harness.}
Inspired by the Nelson--Narens metacognition theory, we
instantiate metacognition as a two-level control loop for language model
reasoning. The meta level monitors the model's reasoning state through
self-reported signals, including pre-solve feeling of knowing (FOK) and
post-solve judgment of learning (JOL), while the object level
performs test-time scaling actions such as solving, retrying, aggregating,
and stopping. A model-specific decision boundary learned from the anchor
diagnosis converts FOK/JOL signals into control decisions, allowing each
model to decide when to trust its current answer and when to allocate
additional computation.
}
    \label{fig:harness}
\end{figure}

Our metacognitive harness is inspired by the Nelson--Narens monitoring--control framework. The
overall workflow is shown in Figure~\ref{fig:harness}. For a model $m$
and problem $x$, the harness first queries a single pre-solve
$\mathrm{FOK}$. Each attempt $k$ then produces a reasoning trace $c_k$,
an answer $a_k$, and a post-solve judgment $\mathrm{JOL}_k$. After each
attempt, the model-specific decision function learned during diagnosis
decides whether to stop or retry:
\[
    g_m(\mathrm{FOK}, \mathrm{JOL}_k)
    \rightarrow \widehat{y}_{m,k} \in \{0,1\},
\]
where $\widehat{y}_{m,k}=1$ means that the current attempt is trusted and
the loop stops, while $\widehat{y}_{m,k}=0$ allocates another
Solve--JOL attempt. If the loop stops after one attempt, its answer is
returned directly. If multiple attempts are generated, a separate
verifier-style aggregator selects one existing attempt by index, and the
selected answer is copied verbatim as the final output.

\paragraph{Context management of reasoning.}
The retry stage is controlled not only by the decision rule
$g_m$, but also by what context is exposed to the next solve attempt.
For attempt $k+1$, we provide a compact metacognitive history
\[
    \mathcal{H}^{\mathrm{reason}}_k
    =
    \{(a_i,\mathrm{JOL}_i,r^{\mathrm{JOL}}_i)\}_{i=1}^{k},
\]
where $r^{\mathrm{JOL}}_i$ is the natural-language reason attached to the
JOL score. We intentionally exclude previous reasoning traces
$\{c_i\}_{i=1}^{k}$ from this context. This design makes retry more
directed than best-of-$N$ sampling, which provides no feedback from prior
attempts, while avoiding the strong anchoring effect of self-refine,
which exposes full previous derivations. In other words, the next attempt
knows what the model was uncertain about, but is not forced to
continue from how the previous solution reasoned.

\paragraph{Context management of aggregation.}
The final aggregation stage uses a different context design. Given
$K$ attempts, the verifier receives
\[
    \mathcal{H}^{\mathrm{agg}}_K
    =
    \{(c_i,a_i)\}_{i=1}^{K},
\]
but not $\mathrm{FOK}$, $\{\mathrm{JOL}_i\}_{i=1}^{K}$, or the JOL
reasons. This separation is intentional. FOK/JOL signals are useful for
controlling whether a problem needs more computation, because they
vary meaningfully across problems. However, within the same problem, JOL
scores across attempts often have low variance and do not reliably
identify the best candidate answer. Therefore, we reserve metacognitive
signals for retry control and hide them during final selection, so the
verifier judges candidate attempts based on their reasoning and answers
rather than from potentially misleading self-confidence cues. Detailed implementations of the harness are provided in Appendix~\ref{app:harness}.

\section{Experiments}

We evaluate whether metacognitive control can improve reasoning without modifying the base model.
All main experiments use Claude Sonnet-4.6 as the fixed solver and change only the inference-time
control procedure. The harness elicits a pre-solve FOK signal and post-solve JOL signals, uses the
diagnosed controller to decide when to retry or stop, and aggregates multiple attempts when needed.

We evaluate on three public reasoning benchmarks: HLE-Verified for expert-level STEM reasoning,
LiveCodeBench v6 for code reasoning, and R-Bench-V for multimodal reasoning. Main results are
reported on the full evaluated benchmark splits, while metacognitive diagnosis uses a separate 100-example anchor set disjoint from the downstream evaluations.
We compare against single-pass inference, vertical test-time scaling methods such as Self-Refine and
budget forcing, and parallel selection methods such as verifier and aggregator. Additional
implementation details, prompts, diagnostic metrics, and ablation settings are provided in Appendix~\ref{app:exp_details}.

\finding{1}{Metacognitive harnessing activates latent reasoning ability.}



\newcommand{\modelcell}[3]{%
  \makecell{#1\\[-0.1em]{\scriptsize #2~#3}}%
}

\newcommand{\bestmodelcell}[3]{%
  \makecell{\textbf{#1}\\[-0.1em]{\scriptsize #2~#3}}%
}

\newcommand{\secondmodelcell}[3]{%
  \makecell{\underline{#1}\\[-0.1em]{\scriptsize #2~#3}}%
}

\newcommand{\harnesscell}[1]{%
  \cellcolor{MCOursGreen}\makecell{#1\\[-0.1em]{\scriptsize \anthropicicon~Sonnet-4.6}}%
}

\newcommand{\bestharnesscell}[1]{%
  \cellcolor{MCOursGreen}\makecell{\textbf{#1}\\[-0.1em]{\scriptsize \anthropicicon~Sonnet-4.6}}%
}

\newcommand{\secondharnesscell}[1]{%
  \cellcolor{MCOursGreen}\makecell{\underline{#1}\\[-0.1em]{\scriptsize \anthropicicon~Sonnet-4.6}}%
}

\newcommand{\gaincell}[1]{%
  \cellcolor{MCGainBlue}\textbf{#1}%
}

\begin{table*}[t]
\centering
\footnotesize
\setlength{\tabcolsep}{3.8pt}
\renewcommand{\arraystretch}{1.15}
\caption{
Results for Finding 1. In Panel A, ``Top-1/2/3'' are single-model leaderboard entries taken from the public benchmark snapshots as of April 2026.
Panel B compares the proposed metacognitive harness against test-time scaling baselines under matched or comparable inference budgets.
}
\label{tab:unified_finding1_finding3}

\begin{tabular*}{\textwidth}{@{\extracolsep{\fill}}llcccccc@{}}
\toprule
\multicolumn{8}{@{}l}{\textbf{A. Main performance against leaderboard models}} \\

\midrule
\rowcolor{MCPanelGray}
Benchmark
& Split
& Top-1
& Top-2
& Top-3
& Pass@1
& Ours
& Gain \\
\midrule

HLE
& \textbf{Gold}
& \secondmodelcell{52.5}{\openaiicon}{GPT-5.2-H}
& \modelcell{50.2}{\anthropicicon}{Opus-4.6}
& \modelcell{48.9}{\googleicon}{Gemini-3-Pro}
& \modelcell{48.0}{\anthropicicon}{Sonnet-4.6}
& \bestharnesscell{60.0}
& \gaincell{+12.0} \\

\midrule

LCB
& \textbf{Overall}
& \secondmodelcell{80.2}{\openaiicon}{o4-mini-H}
& \modelcell{75.8}{\openaiicon}{o3-H}
& \modelcell{74.2}{\openaiicon}{o4-mini-M}
& \modelcell{74.3}{\anthropicicon}{Sonnet-4.6}
& \bestharnesscell{84.3}
& \gaincell{+10.0} \\

LCB
& Easy
& \bestmodelcell{99.1}{\openaiicon}{o4-mini-H}
& \bestmodelcell{99.1}{\openaiicon}{o3-H}
& \bestmodelcell{99.1}{\googleicon}{Gemini-2.5-Pro}
& \secondmodelcell{95.4}{\anthropicicon}{Sonnet-4.6}
& \harnesscell{94.7}
& \gaincell{-0.7} \\

LCB
& Medium
& \bestmodelcell{89.4}{\openaiicon}{o4-mini-H}
& \modelcell{86.5}{\openaiicon}{o4-mini-M}
& \modelcell{84.4}{\openaiicon}{o3-H}
& \modelcell{86.5}{\anthropicicon}{Sonnet-4.6}
& \secondharnesscell{87.6}
& \gaincell{+1.1} \\

LCB
& Hard
& \secondmodelcell{63.5}{\openaiicon}{o4-mini-H}
& \modelcell{57.1}{\openaiicon}{o3-H}
& \modelcell{52.7}{\openaiicon}{o4-mini-M}
& \modelcell{55.0}{\anthropicicon}{Sonnet-4.6}
& \bestharnesscell{74.6}
& \gaincell{+19.6} \\

\midrule

RBV
& \textbf{Overall}
& \modelcell{27.9}{\openaiicon}{GPT-5-mini}
& \modelcell{25.8}{\openaiicon}{o3}
& \modelcell{20.9}{\openaiicon}{o4-mini}
& \secondmodelcell{36.1}{\anthropicicon}{Sonnet-4.6}
& \bestharnesscell{41.1}
& \gaincell{+5.0} \\

RBV
& Math
& \modelcell{48.3}{\openaiicon}{GPT-5-mini}
& \modelcell{48.3}{\openaiicon}{o3}
& \modelcell{43.2}{\openaiicon}{o4-mini}
& \secondmodelcell{53.4}{\anthropicicon}{Sonnet-4.6}
& \bestharnesscell{59.1}
& \gaincell{+5.7} \\

RBV
& Physics
& \modelcell{31.8}{\openaiicon}{GPT-5-mini}
& \modelcell{20.4}{\openaiicon}{o3}
& \modelcell{12.7}{\openaiicon}{o4-mini}
& \secondmodelcell{64.3}{\anthropicicon}{Sonnet-4.6}
& \bestharnesscell{71.3}
& \gaincell{+7.0} \\

RBV
& Counting
& \modelcell{22.6}{\openaiicon}{GPT-5-mini}
& \modelcell{22.1}{\openaiicon}{o3}
& \modelcell{19.0}{\googleicon}{Gemini-2.5-Pro}
& \secondmodelcell{24.6}{\anthropicicon}{Sonnet-4.6}
& \bestharnesscell{29.2}
& \gaincell{+4.6} \\

RBV
& Game
& \secondmodelcell{17.1}{\openaiicon}{o3}
& \modelcell{16.4}{\openaiicon}{GPT-5-mini}
& \modelcell{14.5}{\qwenicon}{Qwen2.5VL-72B}
& \secondmodelcell{17.1}{\anthropicicon}{Sonnet-4.6}
& \bestharnesscell{20.7}
& \gaincell{+3.6} \\

\bottomrule
\end{tabular*}

\vspace{0.75em}

\begin{tabular*}{\textwidth}{@{\extracolsep{\fill}}lccccccc@{}}
\toprule
\multicolumn{8}{@{}l}{\textbf{B. Comparison with test-time scaling baselines on all benchmarks}} \\

\midrule
\rowcolor{MCPanelGray}
Method
& Type
& \makecell{Overall\\Acc.}
& \makecell{Gain}
& \makecell{Oracle\\Acc.}
& \makecell{Avg.\\ Attempts}
& \makecell{Avg. Cost\\per Q.}
& \makecell{Gain\\Rank} \\
\midrule

Sonnet-4.6 Pass@1
& Single-pass
& 48.3
& --
& --
& 1.0
& \$0.29
& -- \\

Self-Refine
& Vertical scaling
& 52.4
& +4.0
& 56.6
& 1.5
& \$0.45
& \#3 \\

Budget Forcing
& Vertical scaling
& 51.5
& +3.2
& 51.5
& 1.0
& \$0.67
& \#4 \\

Verifier Reranking
& Parallel scaling
& 49.8
& +1.5
& 60.5
& 4.0
& \$1.32
& \#5 \\

Aggregator
& Parallel scaling
& 53.7
& +5.3
& 60.5
& 4.0
& \$1.42
& \#2 \\

\rowcolor{MCOursGreen}
Metacognitive Harness
& Harness
& \textbf{56.9}
& \textbf{+8.6}
& \textbf{62.2}
& \textbf{2.4}
& \textbf{\$0.78}
& \textbf{\#1} \\

\bottomrule
\end{tabular*}

\vspace{0.45em}
\begin{minipage}{0.98\textwidth}
\scriptsize
\textit{Notes.}
HLE = HLE-Verified (668 questions); LCB = LiveCodeBench v6 (388 questions); RBV = R-Bench-V (803 questions).
Leaderboard entries are shown as score on the first line and model identity on the second line.
Bold indicates the best score in each row, and underline indicates the second-best score.
H/M denote high/medium reasoning settings when applicable. 

Oracle Acc. counts a question as correct if at least one generated attempt is correct, and therefore measures the reachable upper bound of the explored candidate set rather than a practical inference procedure.
Avg. Attempts reports the realized average number of attempts generated per question.
Avg. Inference Cost / Q. reports the total inference cost divided by the number of evaluated questions. The detailed per-benchmark results of Panel B are provided in Appendix~\ref{app:per_benchmark_results}.
\end{minipage}

\end{table*}

\paragraph{Main result.}
Table~\ref{tab:unified_finding1_finding3} summarizes our main performance results.
Panel A compares the proposed harness with representative top leaderboard models on each benchmark and split.
Across all experiments, Sonnet-4.6 is kept fixed as the base model; the harness only changes the inference-time control procedure.
On the pooled full evaluation set of 1,859 examples, standard Sonnet-4.6 pass@1 achieves 48.3 accuracy, while the metacognitive harness improves it to 56.9, yielding a +8.6 point gain.
The improvement is consistent across domains: the harness improves HLE from 48.0 to 60.0, LiveCodeBench v6 from 74.3 to 84.3, and R-Bench-V from 36.1 to 41.1.
The largest gain appears on the Hard split of LiveCodeBench (+19.6), suggesting that the harness is especially helpful when the questions are hard.

\begin{figure}[t]
  \centering
  \begin{subfigure}{0.32\linewidth}
    \includegraphics[width=\linewidth]{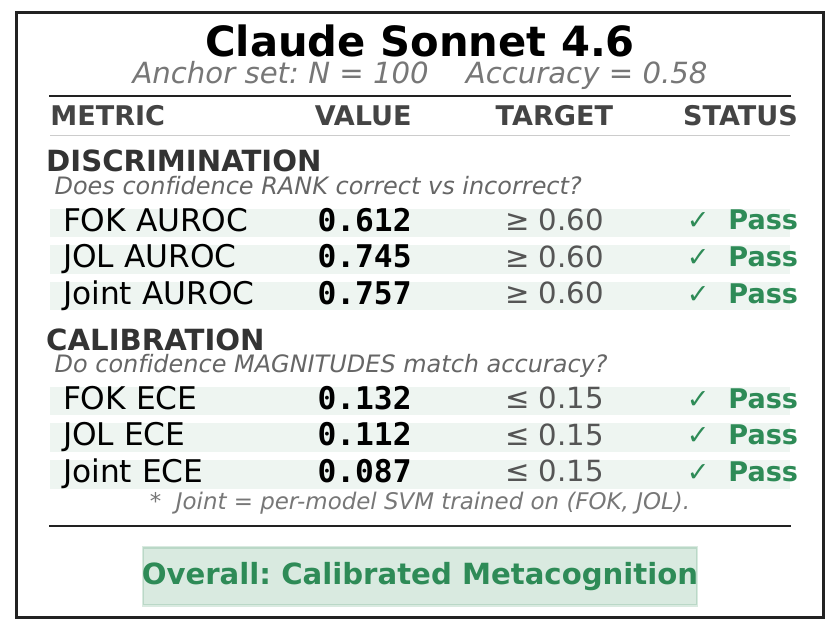}
    \caption{\textsc{Calibrated Metacognition} -- the rubric-passing case.}
  \end{subfigure}\hfill
  \begin{subfigure}{0.32\linewidth}
    \includegraphics[width=\linewidth]{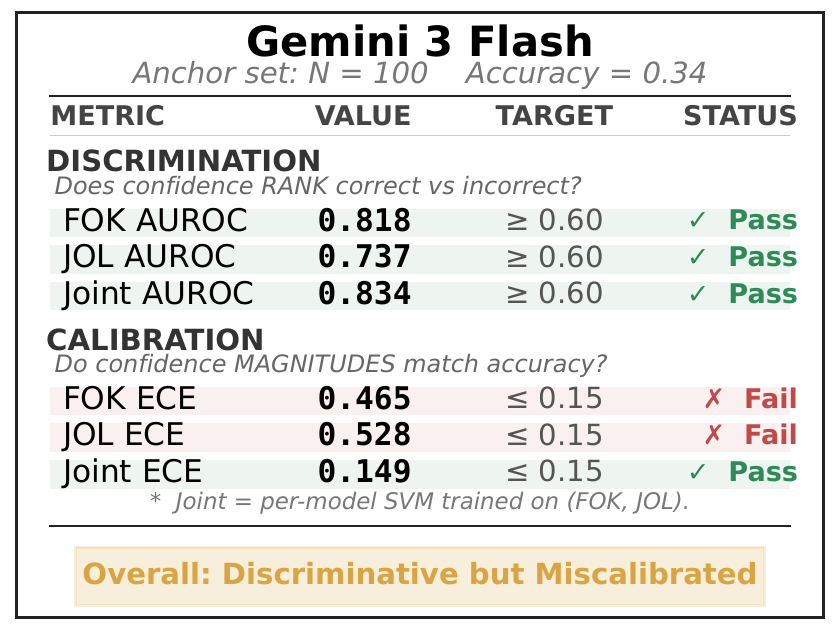}
    \caption{\textsc{Discriminative but Miscalibrated} -- the moderate case.}
  \end{subfigure}\hfill
  \begin{subfigure}{0.32\linewidth}
    \includegraphics[width=\linewidth]{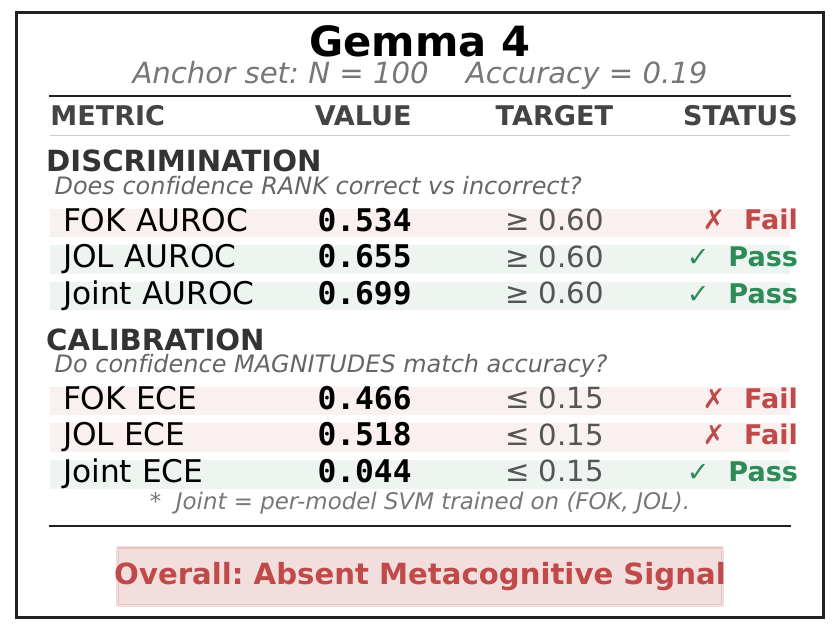}
    \caption{\textsc{Absent Metacognitive Signal} -- the failure case.}
  \end{subfigure}\hfill
  \caption{\textbf{Three diagnosis cards illustrating the verdict
  rubric.} Each card lists six graded rows ($\{$FOK, JOL, Joint$\}
  \times \{$AUROC, ECE$\}$) and a final verdict. Sonnet-4.6 (left) is
  the only model in the panel that passes every row;
  Gemini3-Flash (middle) passes the discrimination rows but fails on calibration ECEs;
  Gemma~4 (right)
  fails discrimination on FOK and calibration on both raw ECEs. The remaining six models fall within the range and are reported in
  Appendix~\ref{app:metacog-diag-extended}.}
  \label{fig:metacog-diag-rubric}
\end{figure}

\begin{figure}[t]
    \centering
    \includegraphics[width=\linewidth]{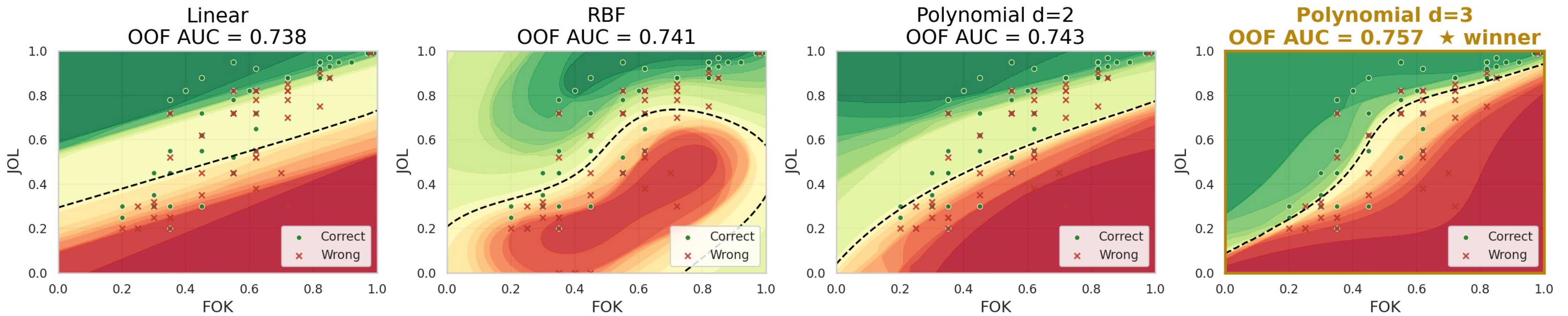}
    \caption{
\textbf{SVM decision function.}
Kernel selection for the joint metacognition classifier on Sonnet-4.6. Each panel shows the decision surface of an SVM trained with StandardScaler and isotonic calibration; titles report out-of-fold AUROC under RepeatedStratifiedKFold (5 splits × 3 repeats). Background color is the predicted P(correct), the dashed line marks the 0.5 boundary.
}
    \label{fig:SVM-kernel-selection}
\end{figure}

\paragraph{Beyond vertical and parallel scaling.}
Panel B further shows that the harness differs from conventional test-time scaling baselines.
We consider two types of baselines.
The first type is vertical scaling, which spends additional computation within a single evolving trajectory.
Self-Refine prompts the model to critique its initial solution and revise or solve again when the answer may be wrong~\citep{madaan2023selfrefine}.
Budget forcing follows the \textsc{s1} protocol: when the model finishes before reaching the target reasoning budget, a continuation cue such as ``Wait'' is appended to encourage further reasoning in the same trajectory~\citep{muennighoff2025s1}.
These methods can improve answers by extending or revising a solution, but they remain conditioned on the previous reasoning context and may therefore be anchored to early mistakes.

The second type is parallel scaling, which generates multiple independent attempts and selects among them after generation.
Verifier reranking applies an external reward model to score a best-of-$N$ candidate set; we use Skywork-Reward-V2 for text-only tasks~\citep{liu2025skyworkrewardv2} and VisualPRM for multimodal tasks~\citep{wang2025visualprm}.
Aggregator selection follows the sample-set aggregation paradigm, where the long-form reasoning traces from multiple parallel attempts are provided to an LLM aggregator, which outputs the final answer~\citep{qi2025learning}.
These methods explore multiple trajectories, but they usually allocate a fixed budget to every example and rely on post-hoc selection.

In contrast, the metacognitive harness performs directed exploration.
It uses the model's own FOK and JOL signals to decide which examples deserve additional attempts, when the current attempt should be trusted, and when generation should stop.
Thus, the harness is not simply longer single-trajectory reasoning or fixed-budget parallel sampling, but adaptive test-time control. 

\paragraph{Exploration oracle and efficiency.}
The oracle accuracy, average-$K$, and cost columns in Panel B further distinguish directed exploration from undirected sampling.
Here, oracle accuracy is not only an upper bound for selection, but also a measure of the exploration policy itself: it asks whether the generated candidate set contains a correct trajectory.
For parallel scaling methods, this oracle reflects what can be found by spending a fixed budget on largely undirected attempts.
In contrast, the harness oracle reflects the candidate set induced by metacognitive control, where FOK and JOL signals guide which examples should receive additional exploration.
Thus, a higher oracle accuracy indicates that the harness is better at reaching useful reasoning trajectories before selection or aggregation even happens.
Average attempts and cost then measure how efficiently this reachable capacity is obtained.
The goal is not to sample more uniformly, but to explore more selectively: easy or confident examples can stop early, while uncertain but promising examples receive additional attempts.

\finding{2}{Metacognitive diagnosis identifies which models are suitable for harnessing.}
\label{sec:finding-metacog}

\paragraph{Diagnosis protocol.}
We use a 100-example anchor set to test whether each model exposes metacognitive signals that are usable for inference-time control.
For each model, we collect a pre-solve FOK score, a post-solve JOL score, and the correctness label of the corresponding attempt.
We evaluate each signal along two axes: discrimination, measured by AUROC, tests whether higher scores rank correct attempts above incorrect ones; calibration, measured by ECE, tests whether confidence magnitudes match empirical correctness.
We also train a lightweight per-model SVM on $(\mathrm{FOK}, \mathrm{JOL})$ to test whether the two signals can be fused into a calibrated joint control score.

\paragraph{Metacognitive signals are common, but harnessable metacognition is rare.}
Figure~\ref{fig:metacog-diag-rubric} summarizes three representative diagnosis outcomes.
Across the nine evaluated models, many models expose discriminative self-monitoring signals, especially through JOL, but far fewer provide confidence values that are calibrated enough to support stopping and retry decisions.
Sonnet-4.6 is the only model in our panel that passes every diagnosis row: its FOK, JOL, and joint SVM signals are all both discriminative and calibrated.
In contrast, Gemini-3-Flash illustrates a discriminative but miscalibrated case, while Gemma-4 illustrates a failure mode where calibration cannot synthesize a reliable signal when the underlying FOK/JOL information is weak.
Thus, a strong pass@1 model is not necessarily harnessable; what matters is whether its self-monitoring signals are predictive and calibratable.

\paragraph{From diagnosis to control.}
The diagnosis is not only descriptive; it determines whether and how a model should be harnessed.
Figure~\ref{fig:SVM-kernel-selection} visualizes the learned SVM decision function for Sonnet-4.6, showing how the controller converts the two-dimensional metacognitive state $(\mathrm{FOK}, \mathrm{JOL})$ into a calibrated probability of correctness and a retry/stop boundary.
For models with discriminative but miscalibrated signals, this learned boundary can convert noisy raw confidence into a more stable control score.
For models with absent or near-random signals, downstream control should not rely on metacognitive outputs.
This diagnosis motivates our use of Sonnet-4.6 as the fixed base model in Finding~1 and explains why metacognitive diagnosis is a necessary step before applying the harness.

\finding{3}{Ablations validate the metacognitive control design.}
\label{sec:finding-ablation}

\begin{table*}[t]
\centering
\scriptsize
\setlength{\tabcolsep}{3pt}
\renewcommand{\arraystretch}{1.10}
\caption{Ablation study on the 100-example set ($N{=}100$, Pass@1${=}$44.0\%, Oracle@4${=}$53.0\%). All variants use Sonnet-4.6 as base solver. Full-system controller is per-model SVM (poly $d{=}3$, $\mathrm{coef}_0{=}1$, isotonic-calibrated, $p_{\rm stop}{=}0.7$) trained on the 100-question anchor set via GroupKFold.}
\label{tab:ablation_harness}
\begin{tabular*}{\textwidth}{@{\extracolsep{\fill}}l@{\hspace{4pt}}l@{\hspace{6pt}}cccp{0.27\textwidth}@{}}
\toprule
\rowcolor{MCPanelGray}
Component & Variant & \makecell{Acc.\,(\%) \\ with 95\% CI} & \makecell{Avg.\\$K$\,$\downarrow$} & \makecell{Early-stop\\hit\,(\%)$\uparrow$} & Takeaway \\
\midrule

Full system
& \textbf{Metacognitive Harness}
& \textbf{47.0\,{\tiny [37,57]}}
& \textbf{2.58}
& \textbf{76.7}
& Full system \\

\midrule
\multirow{2}{*}{Metacognitive signal}
& w/o FOK, JOL-only & 45.0\,{\tiny [35,55]} & 2.83 & 65.8 & No pre-solve solvability estimate \\
& w/o JOL, FOK-only & 42.0\,{\tiny [33,52]} & 3.06 & 67.7 & No post-solve reliability estimate \\
\midrule
\multirow{2}{*}{Controller}
& w/o SVM, $(1{-}\mathrm{JOL})\!\cdot\!\mathrm{FOK}$ & 42.0\,{\tiny [32,52]} & 2.40 & 60.8 & Hand-tuned threshold, no learning \\
& Random retry, matched $K$ & 45.7\,{\tiny [43,48]}
& 2.58
& 42.6
& Same budget, undirected \\
\midrule
\multirow{2}{*}{Aggregation}
& Last answer (all 4 attempts) & 40.0\,{\tiny [31,50]} & 4.00 & -- & Ignores prior attempts \\
& Max-JOL (all 4 attempts) & 44.0\,{\tiny [35,54]} & 4.00 & -- & No answer-level aggregation \\
\midrule
\multirow{2}{*}{Context management}
& No prior metacognitive state & 45.0\,{\tiny [35,55]} & 4.00 & -- & Degenerates to independent sampling \\
& Full previous reasoning context & 40.0\,{\tiny [31,50]} & 3.09 & 68.9 & Risks anchoring to failed trajectories \\
\bottomrule
\end{tabular*}

\vspace{0.4em}
\begin{minipage}{0.98\textwidth}
\scriptsize
\textit{Notes.} \textbf{Confidence intervals.}
Brackets on Acc.\ are 95\% nonparametric bootstrap CIs over the 100-question: we resample the per-record correctness vector with replacement
($B{=}5{,}000$) and report the 2.5\textsuperscript{th}/97.5\textsuperscript{th}
percentiles of the bootstrap-mean distribution.

\textbf{Early-stop hit} is the fraction of records where the controller stopped at $K=1$ and the first attempt was already correct. 

\end{minipage}

\end{table*}

\paragraph{Ablation setup.}
Table~\ref{tab:ablation_harness} analyzes which components of the harness drive the improvement.
We run controlled ablations on a 100-example subset randomly sampled from the testing benchmark data, using Sonnet-4.6 as the fixed base model.
Thus, differences across rows reflect changes in the inference-time control procedure rather than changes in model capacity.
These ablations are used for mechanism analysis; the main results in Finding~1 are evaluated separately on the full benchmark.

\paragraph{Ablation results.}
The ablations show that each part of the harness contributes to the final gain.
Removing either FOK or JOL hurts performance, confirming that pre-solve solvability estimation and post-solve reliability estimation are complementary.
Replacing the learned SVM controller with the best hand-crafted confidence rule we found also degrades accuracy, showing that raw confidence scores are not directly actionable and must be calibrated before being used for retry and stopping decisions.
Random retry and fixed-$K$ variants underperform despite matched or comparable budgets, indicating that the gain does not come simply from more attempts, but from directed exploration.
Aggregation ablations further show that simple rules such as last-answer selection or max-JOL selection are insufficient: the former ignores useful earlier attempts, while the latter over-trusts a single self-rated confidence score.
Finally, context ablations show that retry needs compact metacognitive feedback; using no prior state degenerates toward independent sampling, whereas exposing full previous reasoning traces can anchor later attempts to earlier mistakes.

Overall, the harness is not a prompting trick or a uniform increase in test-time compute.
Its gains come from the interaction of calibrated metacognitive control, directed retry, aggregation, and context management.

\finding{4}{Confidence is useful for allocating effort across problems, but not for selecting among attempts within a problem.}
\begin{figure*}[t]
    \centering

    \begin{subfigure}[t]{0.27\textwidth}
        \centering
        \includegraphics[width=\linewidth]{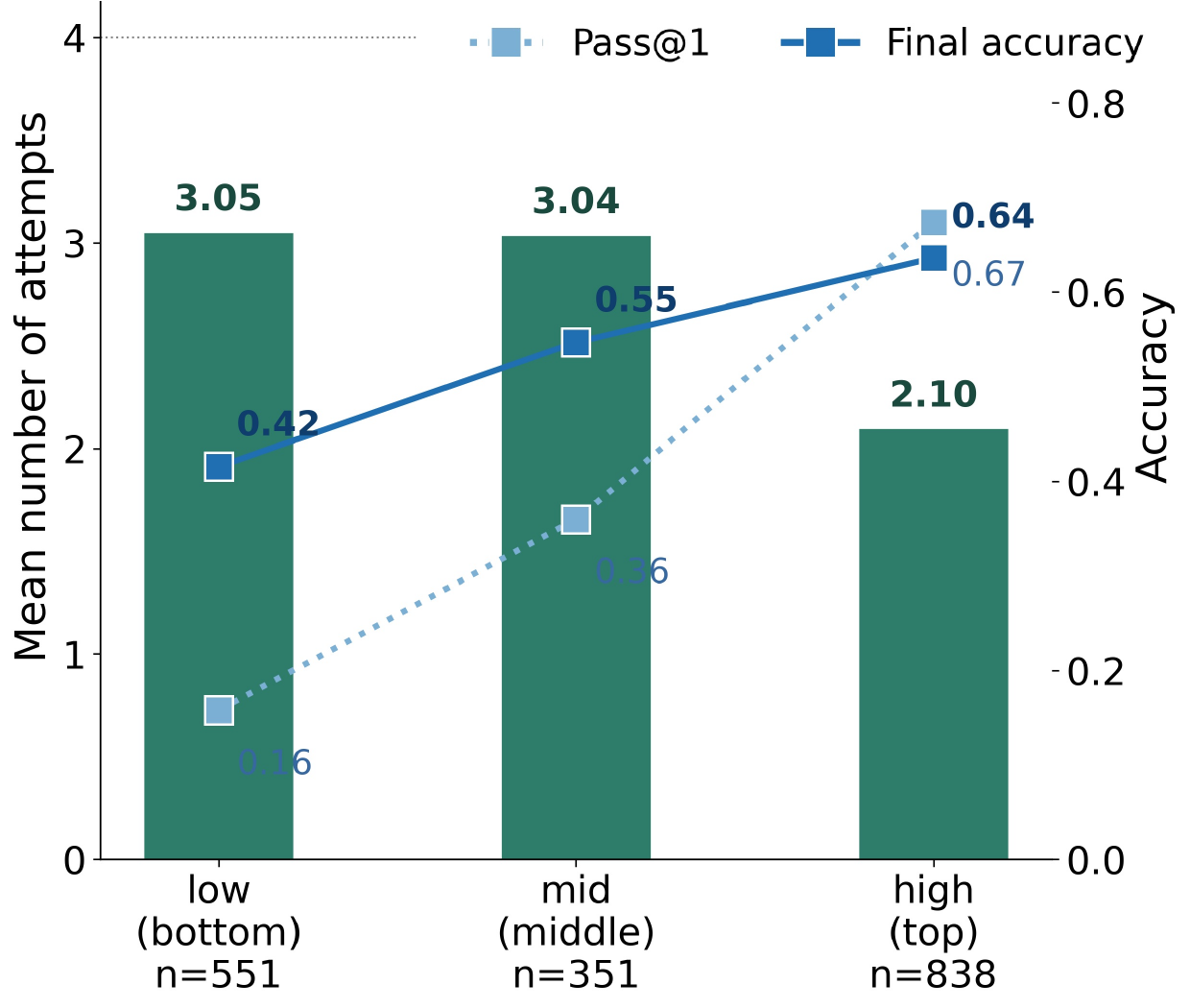}
        \caption{Gains concentrate on low-confidence problems.}
        \label{fig:confidence_gain}
    \end{subfigure}
    \hfill
    \begin{subfigure}[t]{0.30\textwidth}
        \centering
        \includegraphics[width=\linewidth]{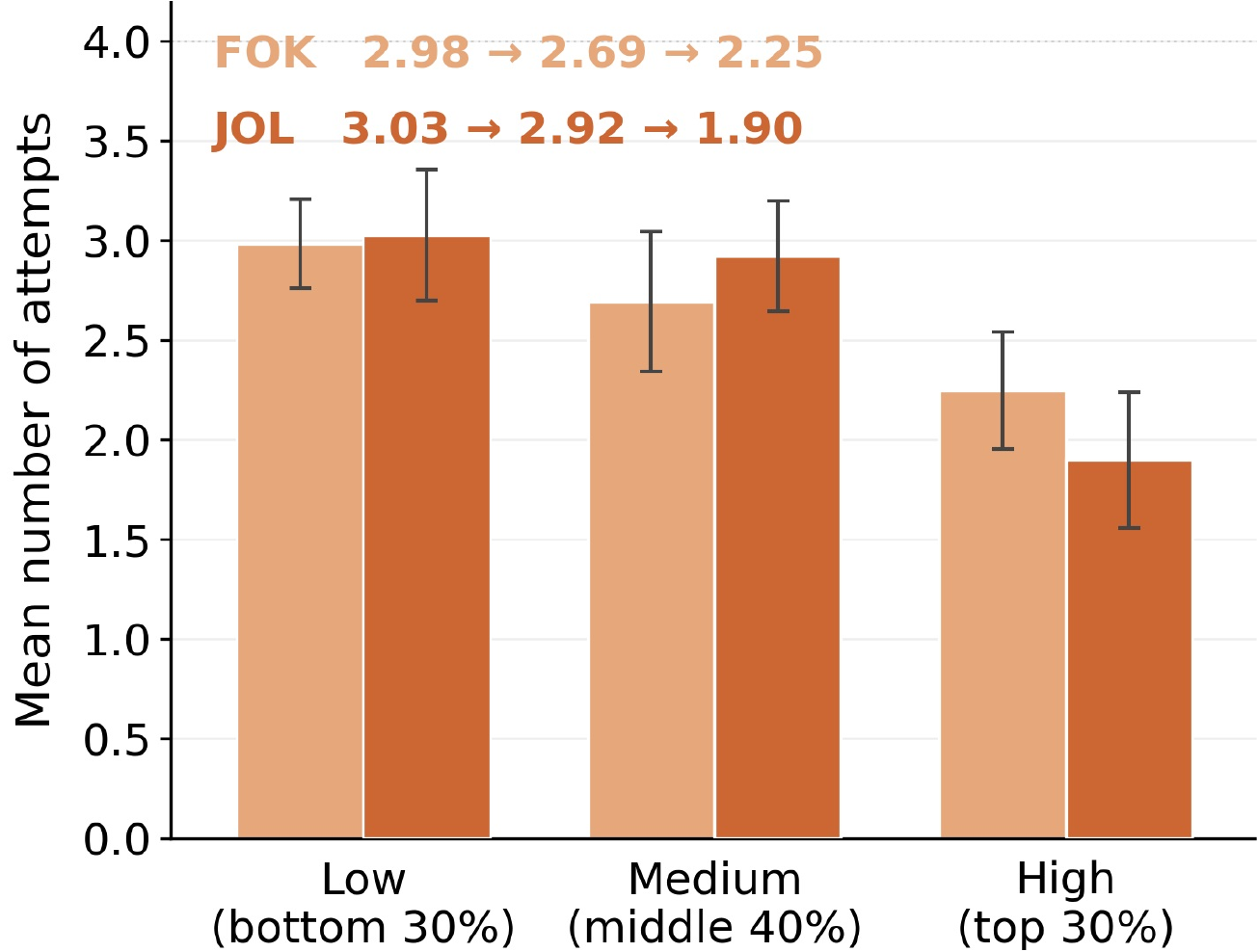}
        \caption{The harness allocates more effort to uncertain problems.}
        \label{fig:adaptive_effort}
    \end{subfigure}
    \hfill
    \begin{subfigure}[t]{0.39\textwidth}
        \centering
        \includegraphics[width=\linewidth]{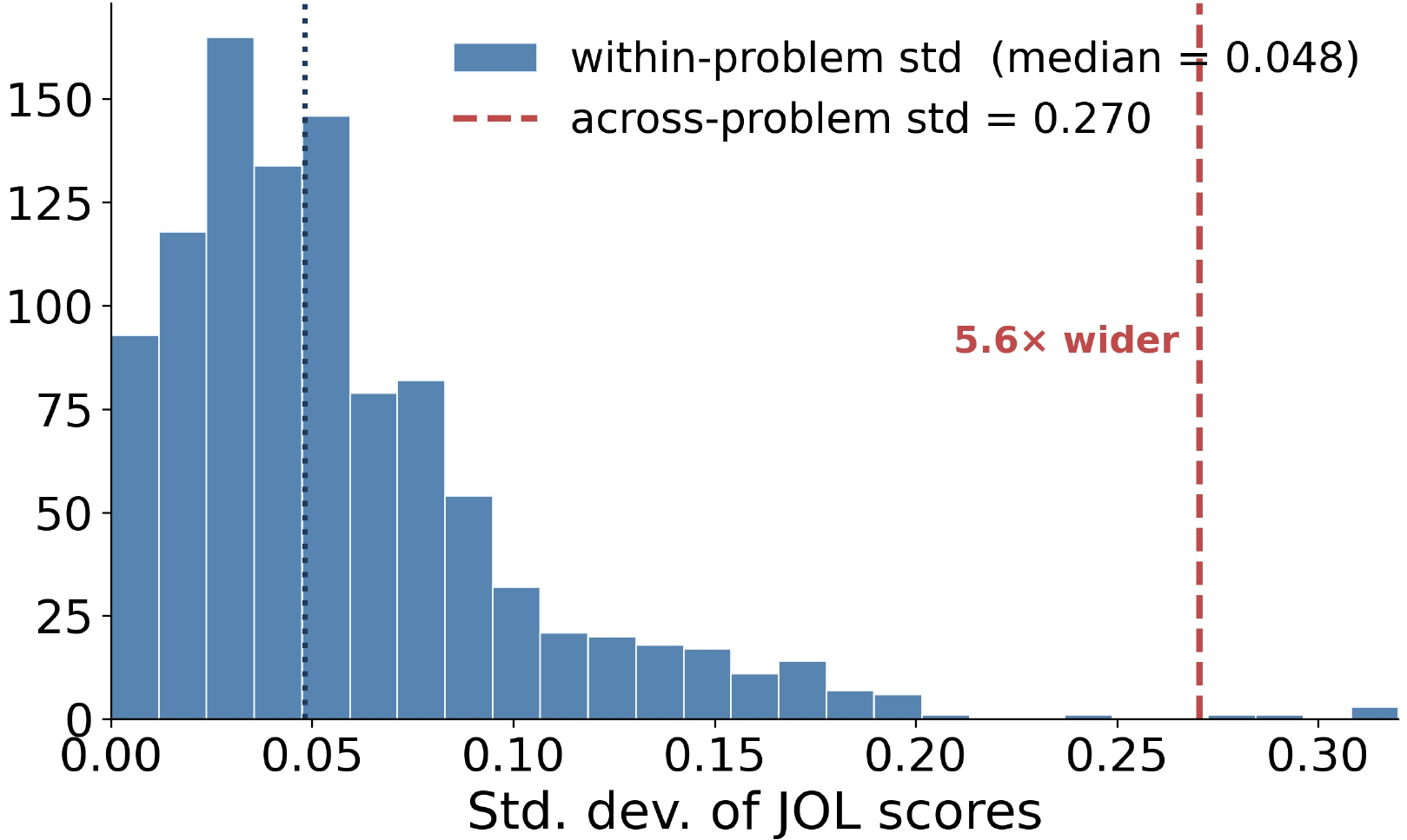}
        \caption{JOL has limited within-problem resolution.}
        \label{fig:JOL_within_problem}
    \end{subfigure}

    \caption{
    \textbf{Discussions of confidence scores.}
    (a) The harness spends more attempts on low-$JOL_1$ problems, where it also yields larger gains.
    (b) After harnessing, effort decreases with confidence for both FOK and JOL.
    (c) JOL varies much more across problems than within the same problem, motivating our choice to use confidence for retry/stop decisions but not for aggregation.
    }
    \label{fig:confidence_control_analysis}
\end{figure*}

\paragraph{Harness gains concentrate on low-confidence problems.}
Figure~\ref{fig:confidence_control_analysis} (a) groups evaluation examples by the first-attempt JOL score $JOL_1$.
The harness allocates nearly the maximum number of attempts to the low- and medium-confidence groups, but substantially fewer attempts to the high-confidence group.
More importantly, the accuracy gain is highly non-uniform:
for low-confidence problems, performance improves from 0.16 at pass@1 to 0.42 after harness control;
for medium-confidence problems, it improves from 0.36 to 0.55;
for high-confidence problems, the harness does not improve performance and instead slightly reduces accuracy, from 0.67 to 0.64.
This shows that the harness is most useful on problems that the model initially regards as uncertain, while providing little benefit on already confident cases.

\paragraph{The harness induces adaptive effort allocation.}
Figure~\ref{fig:confidence_control_analysis} (b) further shows how metacognitive control changes the model's allocation of reasoning effort.
When grouped by confidence, the average number of attempts decreases monotonically from low- to high-confidence examples for both FOK and JOL.
This behavior contrasts with Figure~\ref{fig:harness_engineering} (c), where the unharnessed model does not systematically allocate more effort to low-confidence questions.
Thus, the harness does not merely improve final accuracy; it also converts metacognitive confidence into an explicit policy for adaptive compute allocation.

\paragraph{Confidence should guide retry control, not final aggregation.}
Why, then, do we use FOK/JOL for retry and stop decisions, but hide them during aggregation?
Figure~\ref{fig:confidence_control_analysis} (c) provides the answer.
Across attempts for the same problem, JOL scores have very small variation: the median within-problem standard deviation is only 0.048.
By contrast, the variation of JOL across different problems is much larger, with a standard deviation of 0.270, about 5.6$\times$ wider.
This means that confidence is informative for deciding which problems deserve more computation, but much less informative for distinguishing which attempt is best within the same problem.
Therefore, metacognitive confidence is most useful as a control signal for retry and stopping, while final aggregation should instead rely on the reasoning traces and answers themselves.

\section{Conclusion and Future Work}

We presented a metacognitive harness that turns an LLM's self-monitoring signals into inference-time control.
Rather than treating confidence as a passive diagnostic, the harness uses pre-solve FOK and post-solve JOL signals to decide when to trust, retry, stop, and aggregate.
Across STEM, code, and multimodal reasoning benchmarks, this control layer improves a fixed base model without parameter updates or benchmark-specific fine-tuning.
Our results suggest that strong LLMs may already expose useful self-knowledge, but need an explicit harness to act on it during reasoning.

In future work, we plan to release the code, prompts, and data used for metacognitive diagnosis and controller fitting.
We also plan to package the harness as a reusable interface, such as Model Context Protocol (MCP) and Skill, so that metacognitive control can be easily plugged into existing agent harness systems.
Finally, we will evaluate the approach on agent benchmarks, where retry, stopping, tool use, and context management are central to reliable long-horizon behavior.

\bibliographystyle{unsrt} 
\bibliography{refs} 

\newpage
\appendix

\section*{Appendix}
%

\section{Metacognition Signals Details}
\label{sec:appendix-per-model}

\begin{figure}[http]
    \centering
    \begin{subfigure}{0.32\linewidth}
        \includegraphics[width=\linewidth]{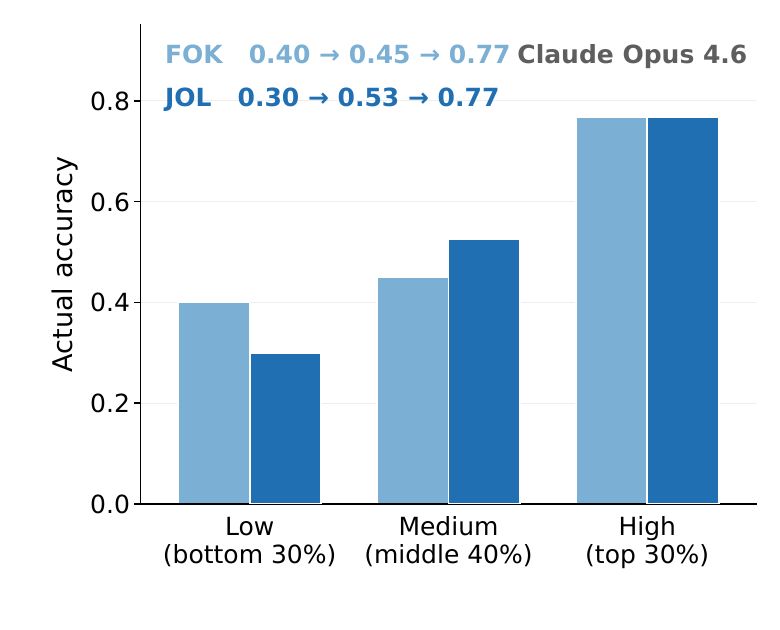}
    \end{subfigure}\hfill
    \begin{subfigure}{0.32\linewidth}
        \includegraphics[width=\linewidth]{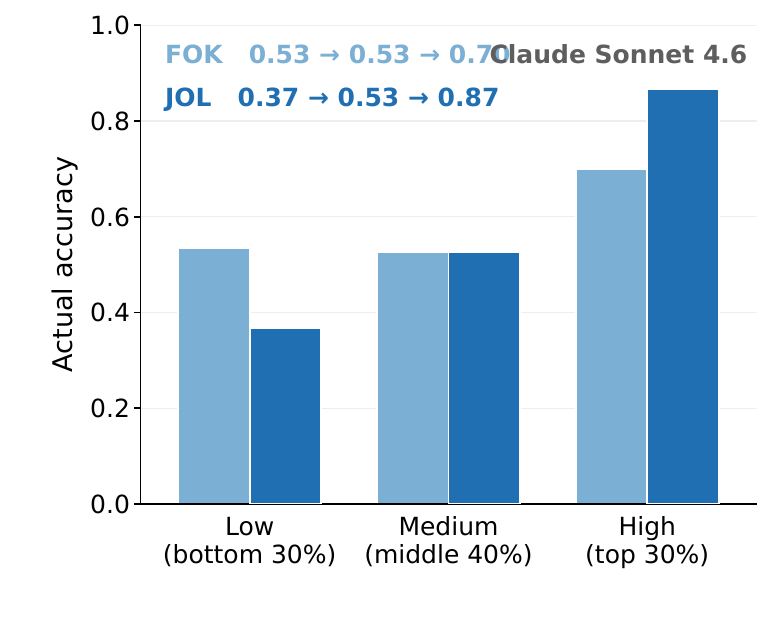}
    \end{subfigure}\hfill
    \begin{subfigure}{0.32\linewidth}
        \includegraphics[width=\linewidth]{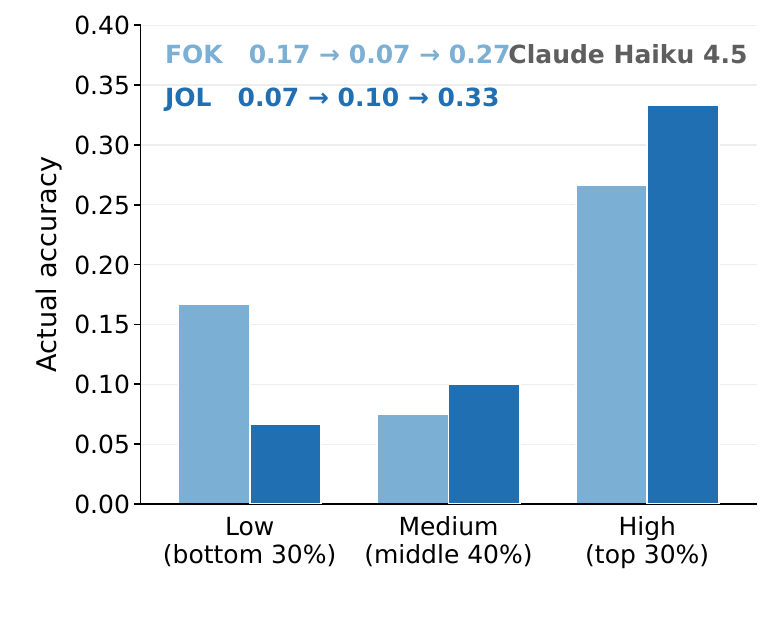}
    \end{subfigure}

    \vspace{0.6em}
    \begin{subfigure}{0.32\linewidth}
        \includegraphics[width=\linewidth]{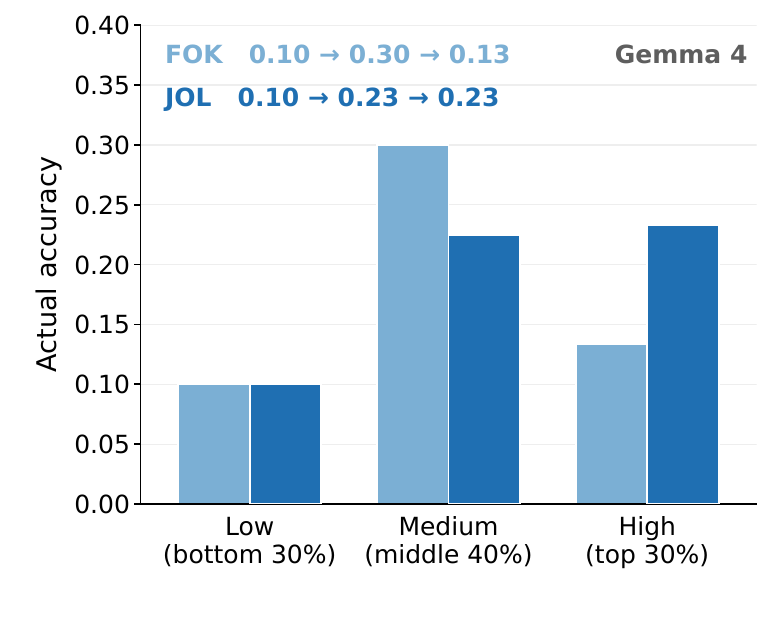}
    \end{subfigure}\hfill
    \begin{subfigure}{0.32\linewidth}
        \includegraphics[width=\linewidth]{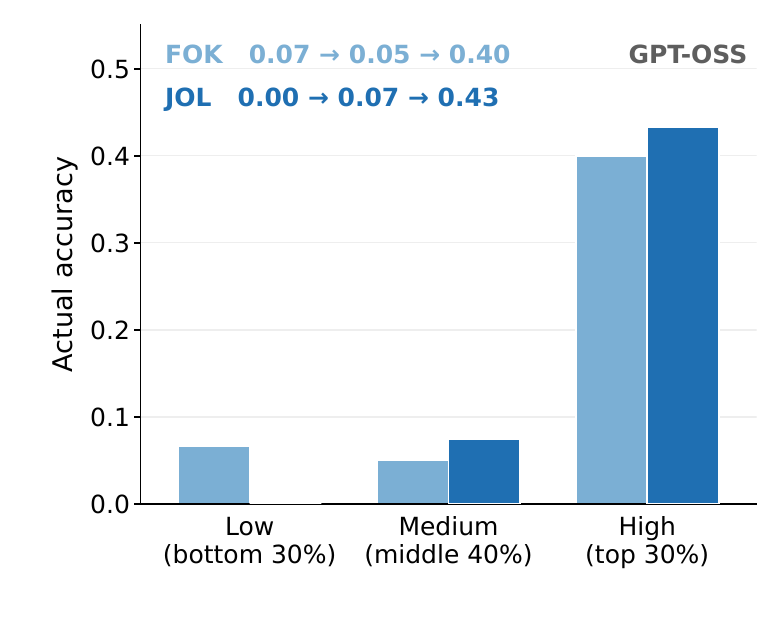}
    \end{subfigure}\hfill
    \begin{subfigure}{0.32\linewidth}
        \includegraphics[width=\linewidth]{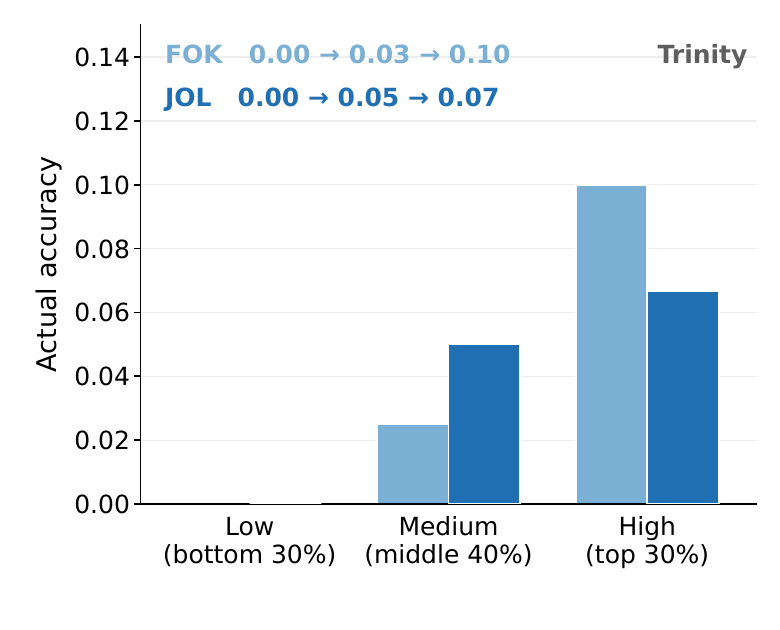}
    \end{subfigure}

    \vspace{0.6em}
    \begin{subfigure}{0.32\linewidth}
        \includegraphics[width=\linewidth]{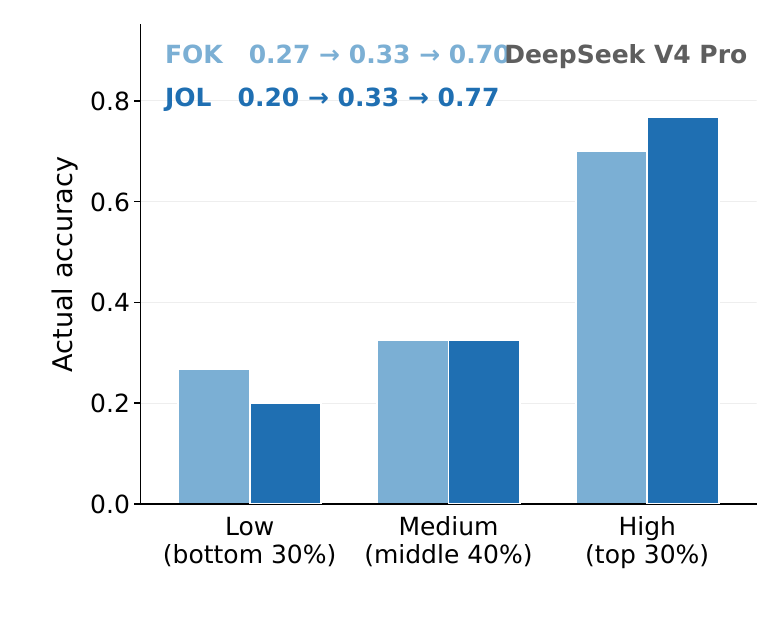}
    \end{subfigure}\hfill
    \begin{subfigure}{0.32\linewidth}
        \includegraphics[width=\linewidth]{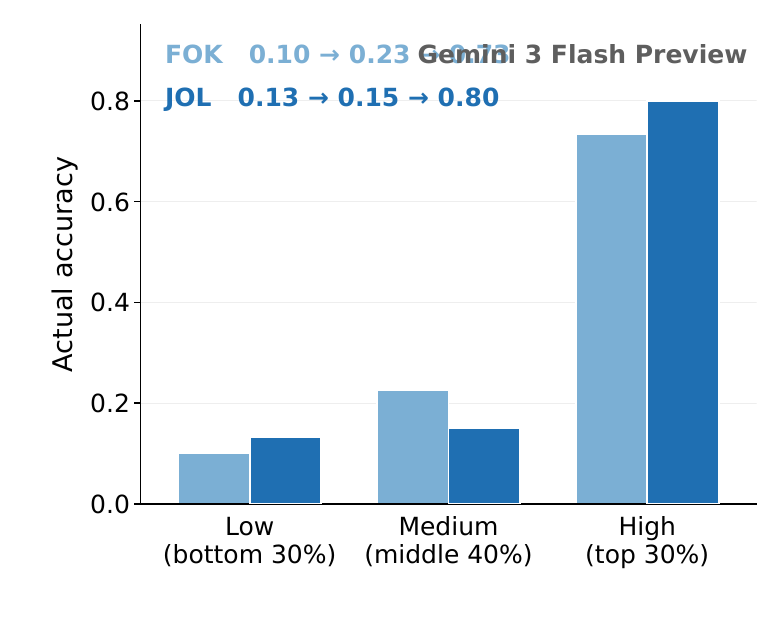}
    \end{subfigure}\hfill
    \begin{subfigure}{0.32\linewidth}
        \includegraphics[width=\linewidth]{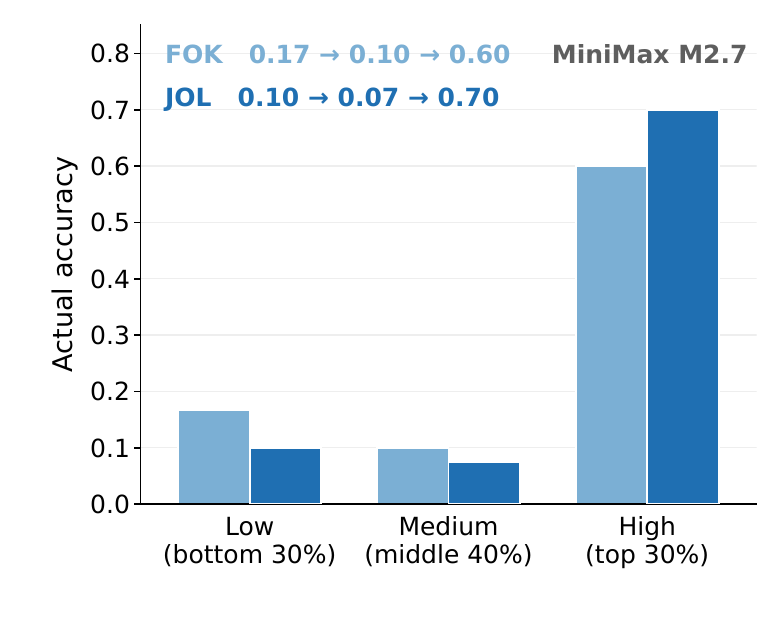}
    \end{subfigure}

    \caption{\textbf{Per-model accuracy by confidence band.}
    Each subplot shows one model's accuracy on the Low (bottom~30\%),
    Medium (middle~40\%), and High (top~30\%) bands for FOK (light blue,
    pre-solve confidence) and JOL (deep blue, post-solve confidence).
    Accuracy generally increases with confidence for most models and signals, though the trend is not strictly monotonic in every case.}
    \label{fig:appendix-panel-b-per-model}
\end{figure}

\begin{figure}[t]
    \centering
    \begin{subfigure}{0.32\linewidth}
        \includegraphics[width=\linewidth]{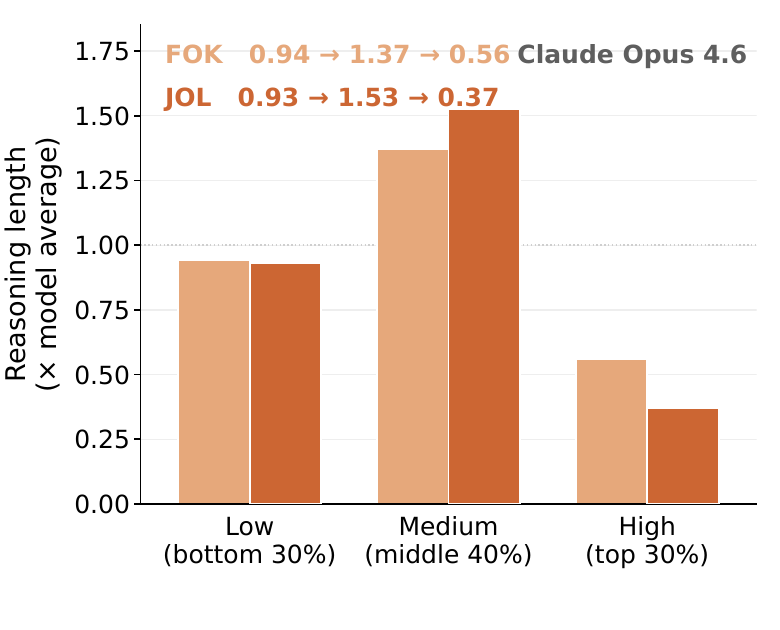}
    \end{subfigure}\hfill
    \begin{subfigure}{0.32\linewidth}
        \includegraphics[width=\linewidth]{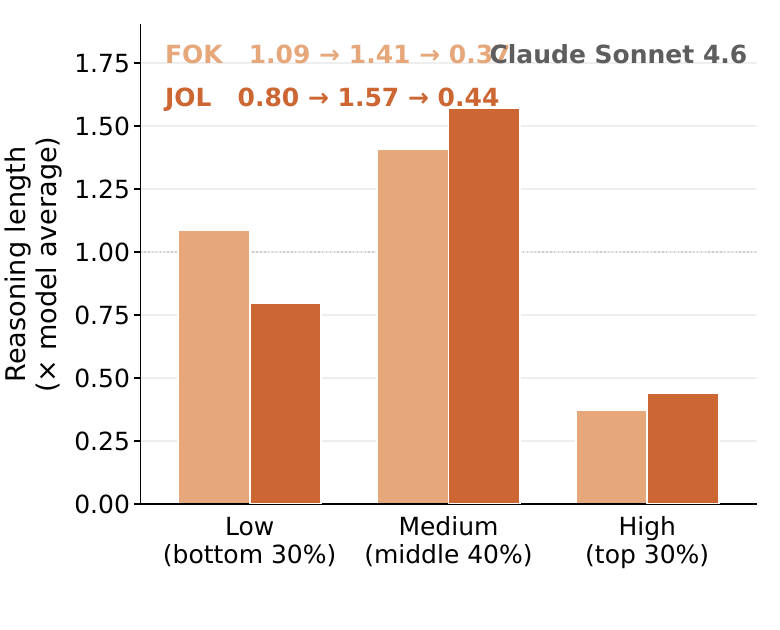}
    \end{subfigure}\hfill
    \begin{subfigure}{0.32\linewidth}
        \includegraphics[width=\linewidth]{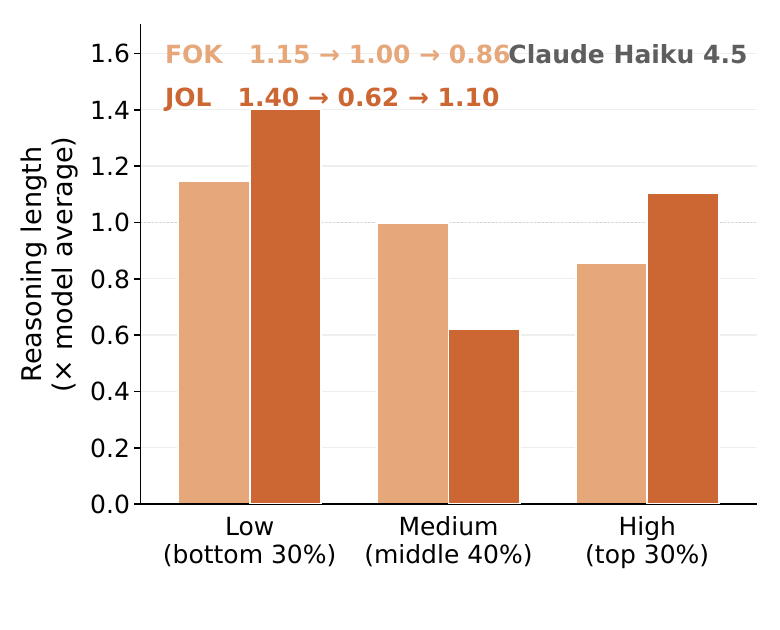}
    \end{subfigure}

    \vspace{0.6em}
    \begin{subfigure}{0.32\linewidth}
        \includegraphics[width=\linewidth]{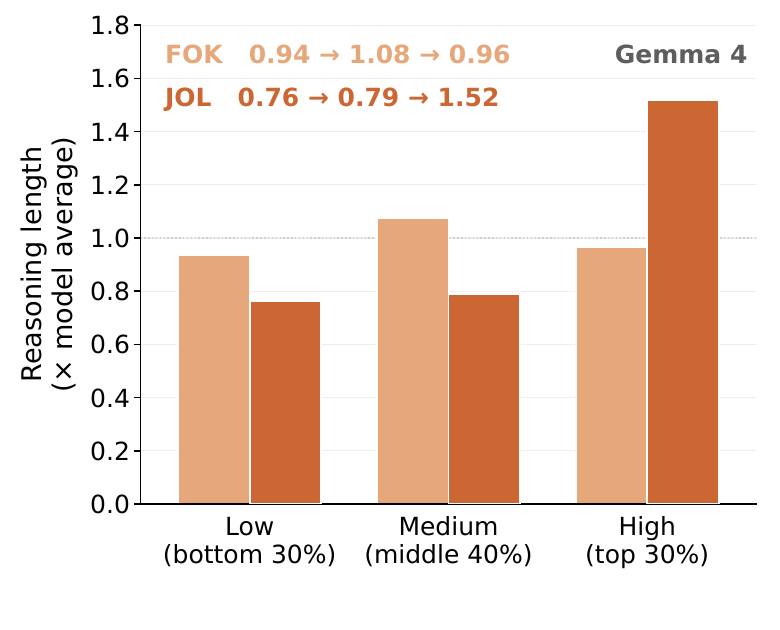}
    \end{subfigure}\hfill
    \begin{subfigure}{0.32\linewidth}
        \includegraphics[width=\linewidth]{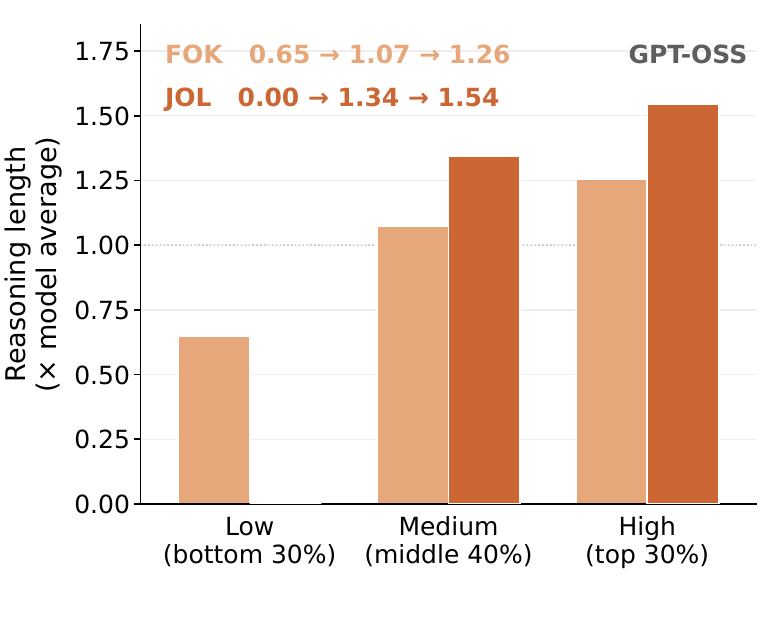}
    \end{subfigure}\hfill
    \begin{subfigure}{0.32\linewidth}
        \includegraphics[width=\linewidth]{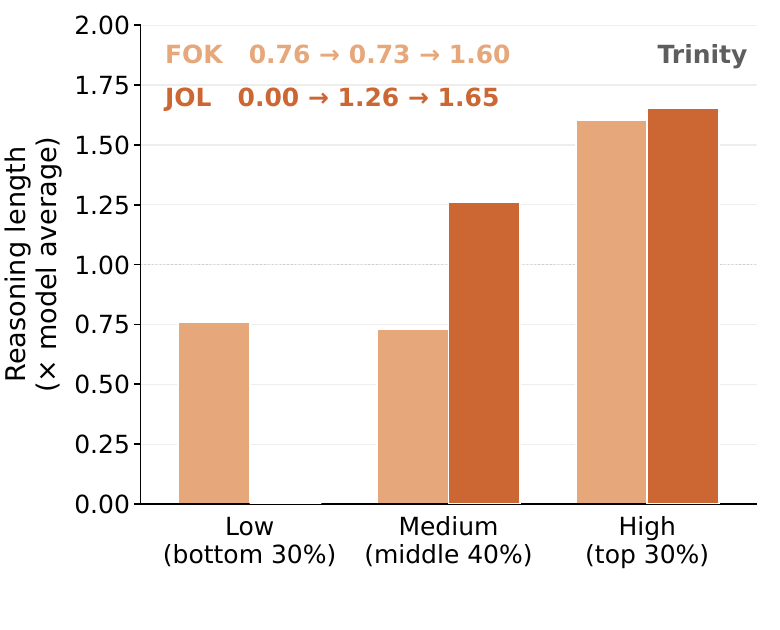}
    \end{subfigure}

    \vspace{0.6em}
    \begin{subfigure}{0.32\linewidth}
        \includegraphics[width=\linewidth]{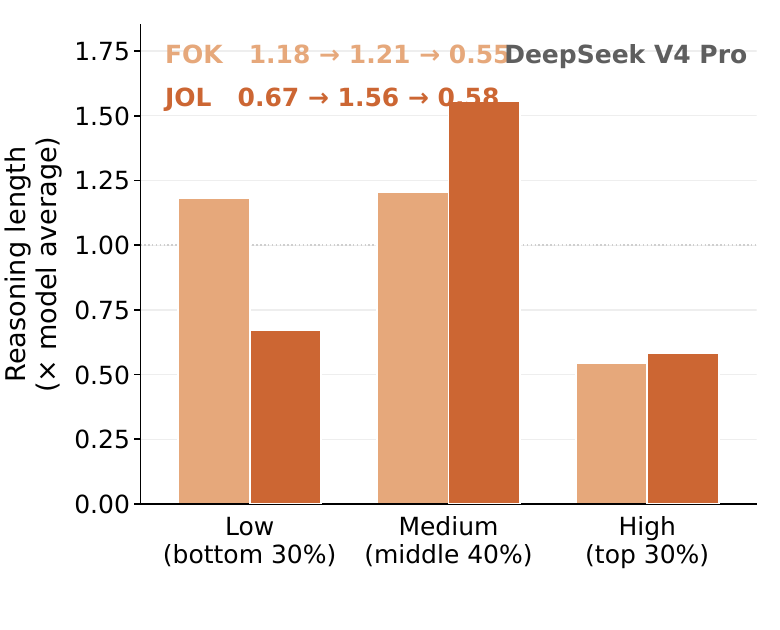}
    \end{subfigure}\hfill
    \begin{subfigure}{0.32\linewidth}
        \includegraphics[width=\linewidth]{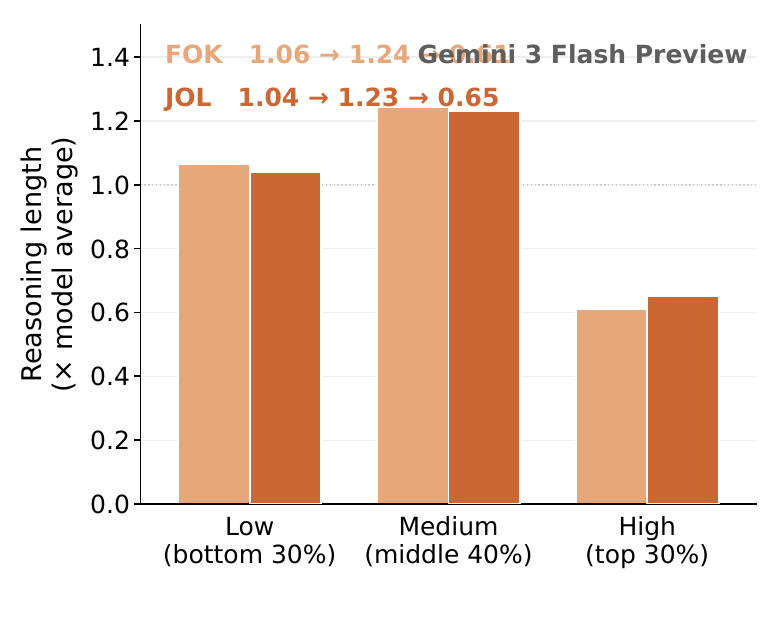}
    \end{subfigure}\hfill
    \begin{subfigure}{0.32\linewidth}
        \includegraphics[width=\linewidth]{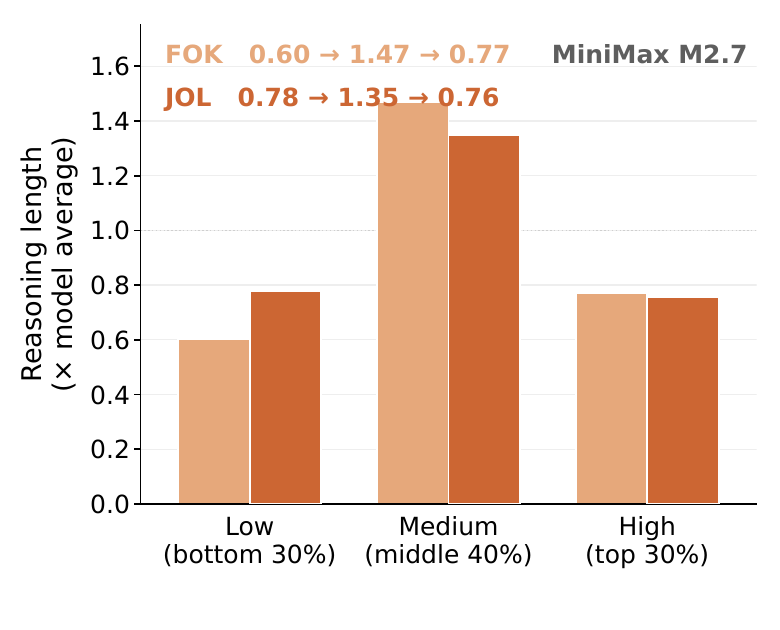}
    \end{subfigure}

    \caption{\textbf{Per-model reasoning length by confidence band.}
    Each subplot shows reasoning length on the Low / Medium / High
    bands for FOK (light orange) and JOL (deep orange), normalised by
    the model's overall mean length (dotted line at $1.0$). The
    direction varies across models — Anthropic models shorten on
    high-confidence items, several open-weight models lengthen — but
    in no case does low confidence consistently buy more reasoning.}
    \label{fig:appendix-panel-c-per-model}
\end{figure}

The main paper reports two cross-model averages: (i) accuracy generally increases with self-reported confidence, and (ii) reasoning length does not follow a consistent confidence-based control pattern. Since averages can mask substantial heterogeneity, we report the per-model breakdown here. For each of the nine models, examples from the 100-item anchor set are sorted independently by FOK (Feeling-of-Knowing, elicited \emph{before} reasoning) and by JOL (Judgment-of-Learning, elicited \emph{after} reasoning). Each signal is split into Low (bottom~30\%), Medium (middle~40\%), and High (top~30\%) bands. We then plot, for each model, the actual accuracy on each band (Figure~\ref{fig:appendix-panel-b-per-model}) and the average reasoning length on each band, normalized by the model's overall mean length so that $1.0$ is the model's own baseline (Figure~\ref{fig:appendix-panel-c-per-model}).

Two broad patterns emerge from the panel. First, accuracy generally increases with confidence for both FOK and JOL, with JOL usually showing a cleaner trend. This indicates that the metacognitive signal is not purely an averaging artifact: In most models, higher self-reported confidence is associated with higher correctness, although the trend is not strictly monotonic in every model and for both signals.

Second, reasoning length varies with confidence in ways that differ substantially across models. The closed-weight Anthropic models often shorten their chains on high-confidence items, especially for FOK and for most JOL panels, whereas several open-weight models show the opposite tendency and lengthen on high-confidence items. These directions therefore depend strongly on the model family rather than reflecting a single shared control strategy. More importantly, the panel does not support a stable picture in which low-confidence cases systematically receive extra verification. Even at the per-model level, current models appear to expose metacognitive signals, but do not reliably use them to allocate reasoning in a controlled and rational way.

\section{Metacognition Diagnosis Details}
\label{app:metacog-diag-extended}
\begin{figure}[http]
  \centering
  \begin{subfigure}{0.48\linewidth}
    \includegraphics[width=\linewidth]{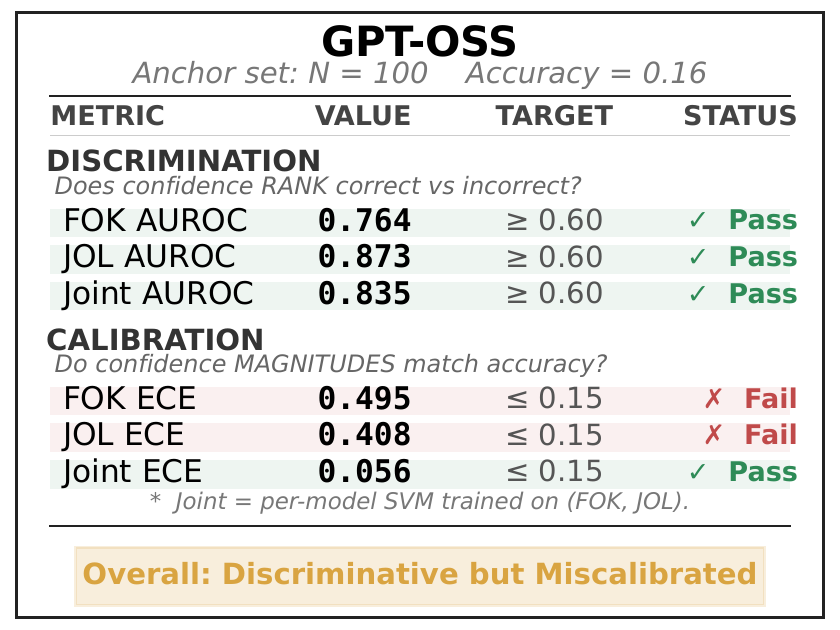}
  \end{subfigure}\hfill
  \begin{subfigure}{0.48\linewidth}
    \includegraphics[width=\linewidth]{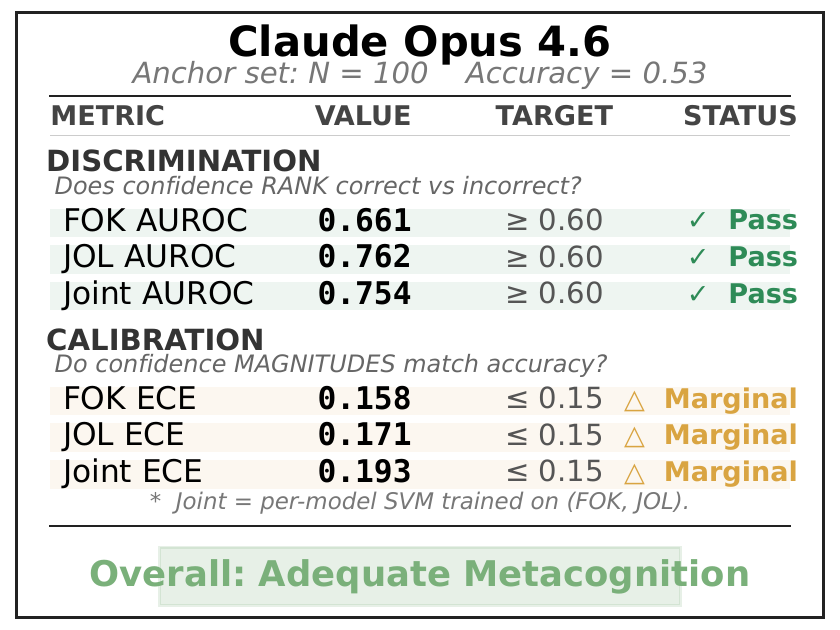}
  \end{subfigure}

  \vspace{0.6em}
  \begin{subfigure}{0.48\linewidth}
    \includegraphics[width=\linewidth]{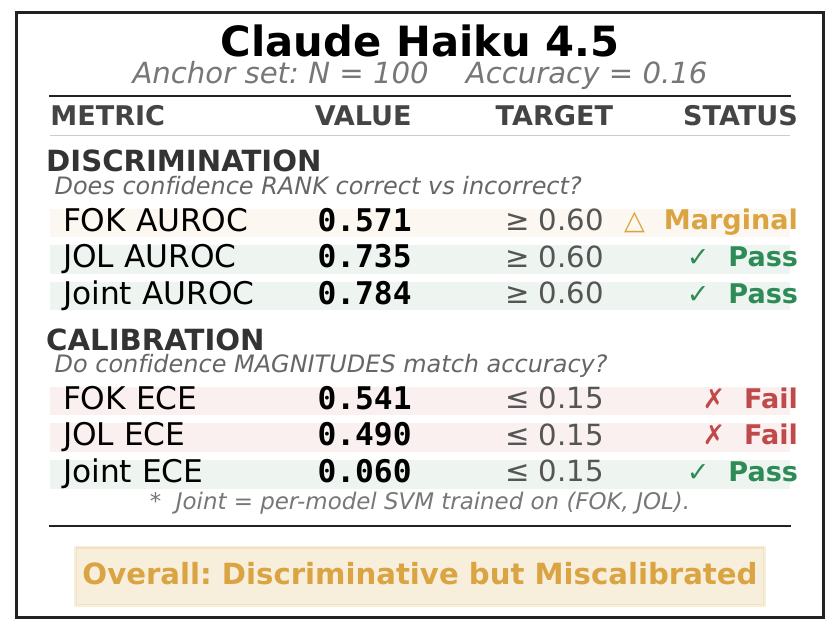}
  \end{subfigure}\hfill
  \begin{subfigure}{0.48\linewidth}
    \includegraphics[width=\linewidth]{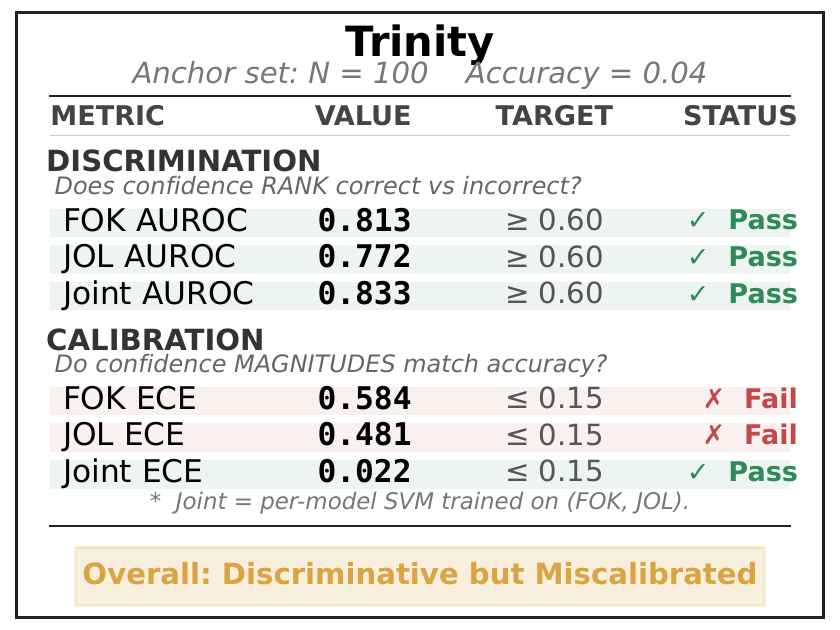}
  \end{subfigure}

  \vspace{0.6em}
  \begin{subfigure}{0.48\linewidth}
    \includegraphics[width=\linewidth]{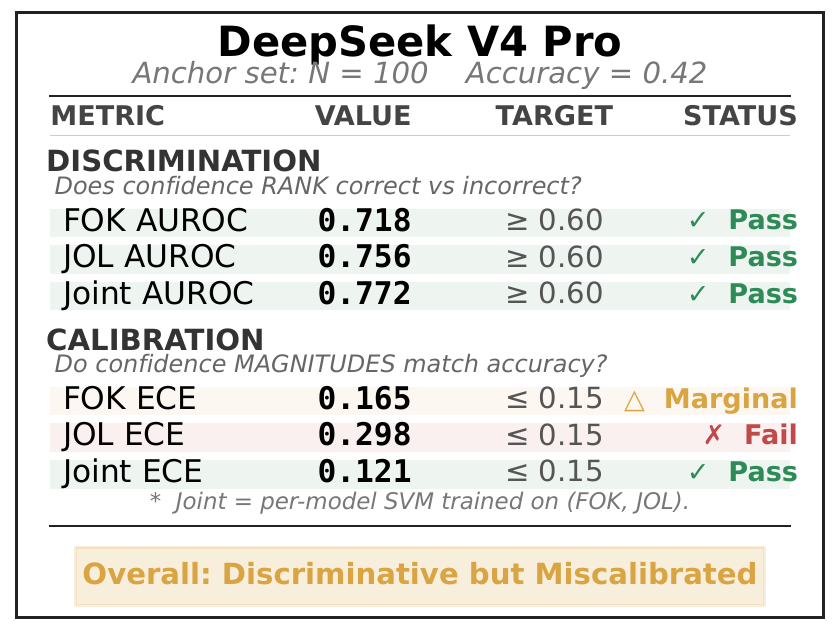}
  \end{subfigure}\hfill
  \hspace*{\fill}
  \begin{subfigure}{0.48\linewidth}
    \includegraphics[width=\linewidth]{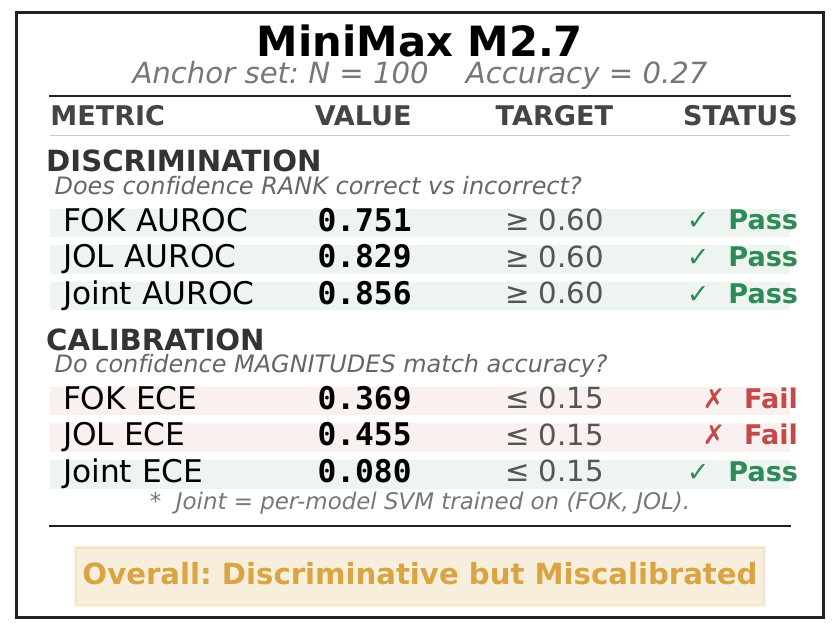}
  \end{subfigure}
  \vspace{0.6em}
  
  \hspace*{\fill}
  \caption{\textbf{Per-model metacognition diagnosis cards
  (extended).} The six models not shown in
  Figure~\ref{fig:metacog-diag-rubric}.Row 1: GPT-OSS, Claude Opus-4.6. Row 2: Claude Haiku 4.5, Trinity. Row 3: DeepSeek V4 Pro, MiniMax M2.7. All five non-Opus models fall into \textsc{Discriminative but Miscalibrated}.}
  \label{fig:metacog-diag-appendix}
\end{figure}

This appendix expands the diagnosis introduced in
Section~\ref{sec:metacog-diag}: the four-level verdict rubric, the
SVM hyperparameter search space, the post-hoc calibration choice, and
diagnosis cards for the six models not shown in the main text.

\paragraph{Diagnosis verdict.}
Each of the six diagnosis rows ($\{$FOK, JOL, Joint$\} \times \{$AUROC,
ECE$\}$) is graded as \textsc{Pass}, \textsc{Marginal}, or
\textsc{Fail} against fixed thresholds (AUROC $\geq 0.60$,
ECE $\leq 0.15$, with \textsc{Fail} at AUROC $< 0.55$ or ECE
$> 0.25$). The six rows are then rolled up into one of four verdicts
that say what the signal can be \emph{used for}:
\begin{itemize}\itemsep0pt
  \item \textsc{Calibrated Metacognition} -- all six rows pass; the
        numerical signal can be used as-is for downstream gating.
  \item \textsc{Adequate Metacognition} -- no row fails outright,
        some sit at the \textsc{Marginal} threshold; rankings and
        probabilities remain usable with mild caution.
  \item \textsc{Discriminative but Miscalibrated} -- the model can
        rank correct vs.\ incorrect but its probability magnitudes
        drift; safe for relative ordering, must be recalibrated before
        being used as a probability.
  \item \textsc{Absent Metacognitive Signal} -- discrimination itself
        is at chance; the signal cannot be trusted.
\end{itemize}

\paragraph{Hyperparameter search.}
For each model we evaluate $67$ estimator configurations crossed with
two post-hoc calibration heads (isotonic regression and Platt's sigmoid
scaling), giving $134$ candidates. The search space covers:
\begin{itemize}\itemsep0pt
  \item \textbf{SVC linear} -- $C \in \{0.01, 0.1, 1, 10, 100\}$.
  \item \textbf{SVC RBF} -- $C \in \{0.1, 1, 10, 100\}\times \gamma \in
        \{\text{scale}, 0.5, 1.0, 2.0, 5.0\}$.
  \item \textbf{SVC polynomial} -- $d \in \{2, 3, 4\}\times \mathrm{coef}_0
        \in \{0, 1, 2\}\times C \in \{0.1, 1, 10\}$.
  \item \textbf{SVC sigmoid} -- $C \in \{0.1, 1, 10\}\times \mathrm{coef}_0
        \in \{0, 1\}$.
  \item \textbf{NuSVC RBF} -- $\nu \in \{0.3, 0.5, 0.7\}$.
  \item \textbf{Logistic regression} -- $\{L_1, L_2\}\times C \in
        \{0.1, 1, 10\}$.
\end{itemize}
Inputs are standardised within each training fold so no test
information leaks through the scaler. All estimators use
\texttt{class\_weight=\,"balanced"}, since accuracy on the anchor set
ranges from $4\%$ (Trinity) to $58\%$ (Sonnet-4.6) and the imbalance
otherwise dominates the SVM hinge loss. We select the configuration
with the highest mean out-of-fold AUROC under $5$-fold $\times
3$-repeat StratifiedKFold, breaking ties by lowest ECE.

\paragraph{Why a post-hoc calibration head.}
The raw SVC \texttt{decision\_function} returns a signed margin on an
arbitrary scale, not a probability, so its ECE is uninformative. We
compared isotonic regression to Platt scaling; isotonic was selected
for $7$ of $9$ models, sigmoid for the other $2$. Across the panel
calibration drops the joint ECE by an order of magnitude relative to
the raw FOK or JOL ECE.

\paragraph{Best configuration per model.}
The winning kernel is not constant across models. Polynomial kernels
($d = 3$, $\mathrm{coef}_0 = 1$) win on Sonnet-4.6, matching the
ablation in our SVM design notes; linear kernels win on Opus-4.6,
Haiku, and Trinity; sigmoid wins on DeepSeek; RBF wins on Gemini and
MiniMax. This is consistent with the FOK--JOL plane having different
curvature in different models, and motivates the per-model search
rather than transferring a single classifier across the panel.

\paragraph{Remaining diagnosis cards.}
Figure~\ref{fig:metacog-diag-appendix} shows the seven models not
shown in Figure~\ref{fig:metacog-diag-rubric}. Six of them are
\textsc{Discriminative but Miscalibrated}; Opus-4.6 is the only one
that reaches \textsc{Adequate Metacognition} (no rows fail, both raw
ECEs sit at the threshold). GPT-OSS is the canonical exemplar of the
discriminative-but-miscalibrated pattern -- JOL AUROC $= 0.87$ but JOL
ECE $= 0.41$ -- with the per-model SVM tightening the joint
probability to ECE $= 0.06$.

\section{Metacognitive Harness Details}
\label{app:harness}

Our pipeline runs five sequential stages on every problem:
(1)~Feeling of Knowing (FOK),
(2)~Solve,
(3)~Judgment of Learning (JOL),
(4)~a retry loop gated by a per-model SVM decision function, and
(5)~a forced-verifier aggregator that \emph{selects} one of the
candidate attempts.

Stages 2 and 3 are conceptually distinct: one generates an answer, and the other evaluates confidence in that answer. For efficiency, however, both are produced in a single SDK call through \texttt{solve\_with\_JOL}. We present them as separate stages because they serve different roles in the pipeline and support different downstream decisions.

Every stage is implemented as a single MCP tool call, so the model's
output is structured JSON rather than free-form text. The system prompt
for each stage forbids any text outside the tool call, which removes a
common failure mode where the model adds an unsolicited ``final
answer'' that contradicts its own tool arguments.

\subsection{Prompting LLMs to Do Self-assessment}

\subsubsection{Stage 1 --- Feeling of Knowing (pre-solve)}

Before the model is allowed to compute anything, it must emit a
gut-feeling assessment via the \texttt{FOK} tool. The system prompt
explicitly forbids any computation, derivation, or partial answer:

\begin{quote}\small
\emph{You are a metacognitive assessment agent.\\
\textbf{FOK (Feeling of Knowing)}: Look at the problem and quickly
assess --- do you feel you know the answer? Give a gut-feeling score
(0--1), a brief domain label, and a short reason explaining your
intuition (e.g.\ ``I recognize this type of problem'' or ``the notation
is unfamiliar''). Do NOT attempt to solve, compute, or derive anything.
No calculations, no steps, no partial answers --- only metacognitive
self-assessment. Your ENTIRE response must be a single tool call.}
\end{quote}

The \texttt{FOK} tool schema requires three fields:
\texttt{domain} (a short topic label),
\texttt{FOK\_score}~$\in [0,1]$ (0 = no idea, 1 = very confident), and
\texttt{FOK\_reason} (a short natural-language justification of the
intuition). The user message is minimal, so as not to leak any solving
cues:

\begin{quote}\small
\texttt{\#\# Problem\textbackslash n\textbackslash n\{problem\}\textbackslash n\textbackslash nGive your Feeling of Knowing. Call \`{}FOK\`{} now.}
\end{quote}

\subsubsection{Stage 2 --- Solve}

The Solve stage produces the reasoning and the candidate answer. It is
restricted to the \texttt{solve\_with\_JOL} tool, whose schema requires
two answer-side fields:
\texttt{reasoning} (a step-by-step chain of thought) and
\texttt{answer} (the final answer string).

The user message carries forward the FOK score and reason as context
so that the solver knows what its own pre-solve intuition was:

\begin{quote}\small
\texttt{\#\# Problem\textbackslash n\textbackslash n\{problem\}\textbackslash n\\
Your FOK score was \{FOK\_score\}.\\
FOK reason: \{FOK\_reason\}\\
Now solve the problem and report your JOL. Call \`{}solve\_with\_JOL\`{} now.}
\end{quote}

Conceptually, this stage is the only one that performs the actual
problem solving --- the reasoning chain produced here is what
downstream stages judge.

\subsubsection{Stage 3 --- Judgment of Learning (post-solve)}

Immediately after producing the answer, the model emits a
\emph{post-hoc} self-assessment of how confident it is. This is
captured by the remaining two fields of the \texttt{solve\_with\_JOL}
tool schema:
\texttt{JOL\_score}~$\in [0,1]$ (0 = pure guess, 1 = certain it is
correct) and \texttt{JOL\_reason} (a brief natural-language explanation
of why this confidence level, e.g.\ ``confident because I verified with
two methods'' or ``unsure because the edge case is tricky'').

Although Solve and JOL are emitted in the same tool call, they are
metacognitively distinct: Solve is object-level reasoning about the
problem, whereas JOL is meta-level reasoning about the reliability of
that reasoning. Forcing the model to produce \texttt{JOL\_reason}
alongside \texttt{JOL\_score} prevents calibration collapse to a default
value (e.g.\ always 0.8) --- the natural-language justification has to
be \emph{specific to this attempt}, which empirically anchors the
score to features of the actual derivation.

\subsubsection{Stage 4 --- Retry under metacognitive gating}

After attempt $k$, the model has produced a post-solve confidence score
$\mathrm{JOL}_k$, while $\mathrm{FOK}$ is obtained once at Stage~1
before any reasoning begins. Rather than using a hand-crafted rule such
as $(1-\mathrm{JOL}_k)\cdot \mathrm{FOK} > \tau$, we use a learned
metacognitive controller
\[
  g_m(\mathrm{FOK}, \mathrm{JOL}_k) \rightarrow \hat{y}_{m,k}\in\{0,1\},
\]
implemented as a per-model SVM fitted on the anchor diagnosis set.
The controller defines a model-specific decision boundary over the
two-dimensional metacognitive state $(\mathrm{FOK}, \mathrm{JOL}_k)$.
If $\hat{y}_{m,k}=1$, the current attempt is trusted and the loop
stops; if $\hat{y}_{m,k}=0$, the harness triggers another
Solve+JOL pass.

This design avoids imposing a universal threshold on raw confidence
scores, whose semantics and calibration can differ substantially across
models. Instead, the SVM converts the model's self-reported FOK and JOL
signals into a calibrated retry/stop decision that is adapted to the
metacognitive behavior of the specific base model.

When a retry is triggered, the next solve attempt receives a compact
metacognitive history rather than the full previous reasoning traces.
Concretely, we pass forward prior answers together with their JOL scores
and JOL reasons, and instruct the model to try a different method while
addressing the concerns surfaced by earlier self-evaluations:

\begin{quote}\small
\texttt{\#\# Previous Attempts\\
The following are your previous attempts at this problem. Each attempt
includes the answer, confidence (JOL), and the reason for that
confidence level. You should try a DIFFERENT approach and address the
concerns raised in previous JOL reasons.\\
\#\#\# Attempt \#1\\
- **Answer**: \{answer\}\\
- **JOL Score**: \{JOL\_score\}\\
- **JOL Reason**: \{JOL\_reason\}\\
\ldots\\
Now try again with a different method. Call \`{}solve\_with\_JOL\`{} now.}
\end{quote}

Critically, only \texttt{answer}, \texttt{JOL\_score}, and
\texttt{JOL\_reason} from prior attempts are exposed to the next round;
the full \texttt{reasoning} traces are intentionally excluded.
This keeps the retry context compact and reduces anchoring to an earlier
failed derivation, while still providing enough metacognitive feedback
to make the next attempt more directed than independent sampling.
Each retry then re-runs Stage~2 (Solve) and Stage~3 (JOL).

\subsubsection{Stage 5 --- Forced-aggregator selection}

After the retry loop ends, the model has $N{\geq}1$ attempts. If
$N{=}1$, the resulting answer is used directly. Otherwise we invoke a separate
agent in a \emph{verifier} role. Unlike a free-form synthesizer (which
empirically tends to ``re-solve'' the problem and emit a novel, often
wrong, answer not present in any attempt), our aggregator is forced to
\emph{select} one of the existing attempts by index. The system prompt
is:

\begin{quote}\small
\emph{You are a meticulous answer judge. You will see a problem (and
image, if any) along with several candidate solution attempts. Each
attempt has its own reasoning and proposed answer.}

\emph{YOUR TASK: select the SINGLE BEST attempt by its index. You are
NOT allowed to produce a new answer; you are only choosing which
existing attempt is most likely correct.}

\emph{Selection criteria (in order of importance):}
\begin{enumerate}
  \item \emph{Internal consistency of reasoning --- does each step follow from the previous?}
  \item \emph{Mathematical / logical validity --- are computations and deductions correct?}
  \item \emph{Alignment with the problem statement and image (if any).}
  \item \emph{Robustness --- does the conclusion hold up under scrutiny?}
\end{enumerate}

\emph{DO NOT: invent a new answer not present in any attempt; re-solve
the problem from scratch; be biased by attempt order --- the candidates
are presented in random order.}
\end{quote}

The model is given access to a single tool, \texttt{select\_attempt},
whose schema accepts an integer \texttt{selected\_index} $\in
\{1,\dots,N\}$ and a 2--3 sentence \texttt{justification}. The
\texttt{minimum}/\texttt{maximum} bounds in the JSON schema make
out-of-range selections invalid at the tool layer. The user message
shuffles the attempts with a random permutation $\pi$ to remove any
positional bias:

\begin{quote}\small
\texttt{\#\# Problem\textbackslash n\textbackslash n\{problem\}\textbackslash n\\
\#\# Candidate Attempts (\{N\} total, shown in random order)\\
\#\#\# Attempt 1\\
**Answer:** \{attempts[\(\pi(1)\)].answer\}\\
**Reasoning:**\\
\{attempts[\(\pi(1)\)].reasoning\}\\
\ldots\\
Examine the attempts above and call the \`{}select\_attempt\`{} tool with
the index (1..\{N\}) of the single best attempt.}
\end{quote}

The selected shown-index is mapped back through $\pi^{-1}$ to recover
the original attempt, whose \texttt{answer} field is used \emph{verbatim}
as the system's final answer. Up to three retries are allowed in case
the model fails to call \texttt{select\_attempt}; on persistent
failure, the most recent attempt is used as a fallback.

\subsection{Context of Each Stage}

\begin{table}[h]
\centering
\small
\setlength{\tabcolsep}{4pt}
\caption{Information flow across the five stages. Each row lists what
is visible to the model in that stage. Stages 2 and 3 are emitted in a
single tool call but are conceptually distinct: Stage 2 is object-level
solving, Stage 3 is meta-level rating of that solving. Stage 4 (retry)
deliberately drops previous reasoning chains and only exposes
\texttt{(answer, JOL\_score, JOL\_reason)} so the model is encouraged to
take a fresh approach. Stage 5 is run as a fresh agent --- it does
\emph{not} inherit the FOK/JOL signals or the chain history of the
solver, only the problem and the candidate attempts in randomised
order.\newline}
\begin{tabular}{@{}lp{3.4cm}p{4.5cm}p{3.4cm}@{}}
\toprule
\textbf{Stage} & \textbf{System role} & \textbf{User context (in)} & \textbf{Output (out)} \\
\midrule
1.\ FOK            & Metacognitive assessor; forbidden from solving & \{problem, image\} & \{domain, FOK, FOK\_reason\} \\
2.\ Solve          & Same agent, now solving                        & \{problem, image, FOK, FOK\_reason\} & \{reasoning, answer\} \\
3.\ JOL            & Same agent, post-hoc self-rating               & \{problem, image, FOK, FOK\_reason, own reasoning, own answer\} (all in-call) & \{JOL\_score, JOL\_reason\} \\
4.\ Retry $k$      & Same agent, instructed to try a different method & \{problem, image, FOK, FOK\_reason, history of (answer, JOL, JOL\_reason)\} \emph{without} previous reasoning chains & new \{reasoning, answer, JOL\_score, JOL\_reason\} \\
5.\ Select         & Independent judge agent; forbidden from producing new answers & \{problem, image, shuffled list of (answer, reasoning) for all attempts\} & \{selected\_index, justification\} \\
\bottomrule
\end{tabular}

\label{tab:stage-context}
\end{table}

A few design choices are worth highlighting:

\begin{itemize}
  \item \textbf{No reasoning leakage from Stage 1 to Stage 2.} The
  \texttt{FOK\_reason} from Stage 1 is short and intuition-level by
  construction (the system prompt forbids derivations), so feeding it
  into Stage 2 anchors the solver to its own gut feeling without
  short-circuiting the reasoning.

  \item \textbf{Solve/JOL are co-emitted but functionally split.}
  Stages 2 and 3 share a tool call for efficiency, but their outputs
  are consumed by different downstream stages: \texttt{reasoning} +
  \texttt{answer} flow into Stage 5 (verifier), while
  \texttt{JOL\_score} + \texttt{JOL\_reason} flow into Stage 4 (retry
  gate). Treating them as one tool call also forces the model to
  commit to its answer \emph{before} naming a confidence number, which
  prevents post-hoc rationalisation.

  \item \textbf{Retry context is intentionally lossy.} In Stage 4 we
  pass only \texttt{(answer, JOL\_score, JOL\_reason)} of each prior
  attempt --- the full reasoning chain is dropped. This keeps the
  prompt short and prevents the model from re-using the same flawed
  derivation. The instruction ``try a DIFFERENT approach and address
  the concerns raised in previous JOL reasons'' makes the prior
  \texttt{JOL\_reason} act as targeted negative feedback.

  \item \textbf{Stage 5 is a clean-slate verifier.} The aggregator does
  \emph{not} see the solver's FOK or JOL scores, only the candidate
  reasonings. Confidence signals from the solver are correlated with
  surface features (length, fluency) more than with correctness, so we
  want the judge to evaluate each attempt on its reasoning merits. The
  random permutation $\pi$ further removes positional bias.

  \item \textbf{Forced selection vs.\ free-form synthesis.} A free-form
  summarizer frequently produces an answer absent from any candidate;
  the \texttt{select\_attempt} tool eliminates this failure mode by
  construction, since the only valid output is an integer index. The
  selected attempt's \texttt{answer} field is then copied verbatim ---
  no paraphrase, no re-derivation.
\end{itemize}

\subsection{Connection to the Nelson--Narens Metamemory Framework}
\label{app:nelson-narens}

\begin{figure}[t]
    \centering
    \includegraphics[width=\linewidth]{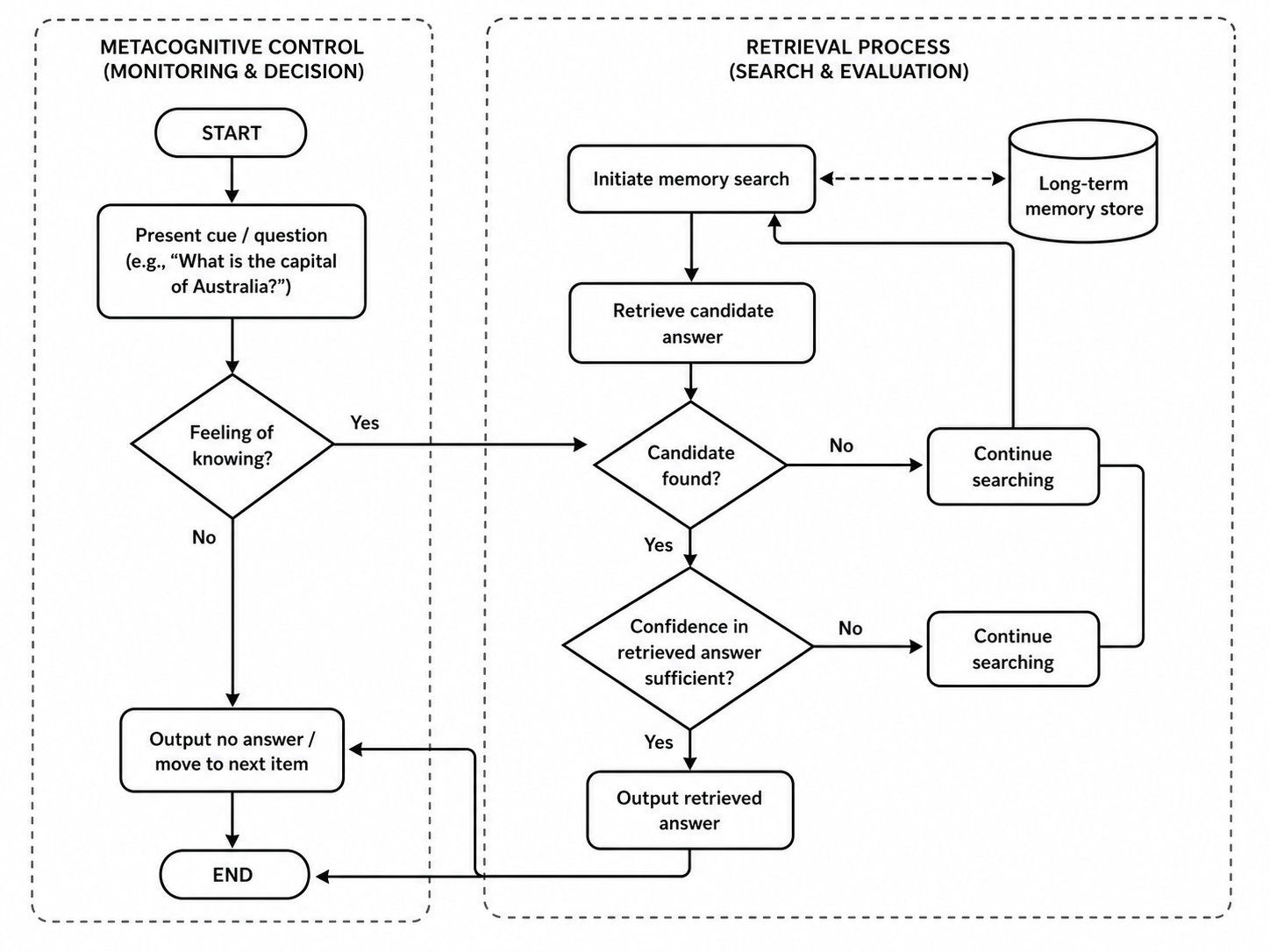}
    \caption{
\textbf{Retrieval-stage metamemory in the Nelson--Narens framework.}
A modern schematic of the retrieval-stage monitoring--control process
described by Nelson and Narens~\citep{nelson1990metamemory}. A
preliminary feeling-of-knowing judgment gates whether search should be
initiated; candidate retrieval and confidence evaluation then determine
whether to output an answer, continue searching, or terminate with no
answer. We use this cognitive framework as the conceptual basis for our
metacognitive harness, where FOK/JOL signals control retry and stopping
decisions during language-model reasoning.
}
\label{fig:nelson-narens-retrieval}
\end{figure}

Our harness is inspired by the monitoring--control view of metamemory
proposed by Nelson and Narens~\citep{nelson1990metamemory}. In this
framework, cognition is organized into two interacting levels: an
\emph{object level}, where the primary cognitive process is carried out,
and a \emph{meta level}, which maintains a model of the object-level
process. The two levels interact through two directional relations:
\emph{monitoring}, where information from the object level updates the
meta level, and \emph{control}, where the meta level modifies the
object-level process by initiating, continuing, or terminating an action.

This distinction is particularly relevant to retrieval. Nelson and Narens
describe self-directed memory retrieval as a process in which a rapid
preliminary feeling-of-knowing (FOK) judgment can determine whether a
person initiates a search at all. If no candidate answer is found, the
person may decide whether to continue searching based on the remaining
FOK signal and the cost or reward of further search. If a candidate answer
is retrieved, a separate confidence judgment determines whether the answer
should be produced or whether search should continue. Thus, retrieval is not
only an object-level search process; it is regulated by meta-level signals
that determine when to search, when to stop, and when to trust a retrieved
answer.

We adapt this monitoring--control structure to language-model reasoning.
In our setting, the object level corresponds to solve attempts, while the
meta level corresponds to self-assessment signals and the learned control
rule. FOK is queried before solving and provides a pre-solve monitoring
signal; JOL is queried after each attempt and provides post-solve feedback
about the current answer. The harness then uses these signals to control
test-time computation: it decides whether to stop, retry with a new
solution path, or aggregate multiple candidate attempts. This is not a
literal model of human memory retrieval, but an operational translation of
the Nelson--Narens principle that metacognitive monitoring becomes useful
only when it is coupled to control.

Figure~\ref{fig:nelson-narens-retrieval} redraws the retrieval-stage
metamemory process in a modern schematic form. The figure emphasizes the
same conceptual loop that motivates our method: a meta-level judgment
first gates whether search should begin; retrieval then produces either no
candidate or a candidate answer; subsequent metacognitive evaluations
decide whether to continue searching, output an answer, or terminate with
no answer. Our harness follows the same high-level logic, replacing human
memory search with test-time reasoning attempts and replacing hand-coded
retrieval thresholds with model-specific decision boundaries learned from
anchor diagnoses.

\section{Model Details}
\label{app:model_details}

\paragraph{Base model.}
All pass@1 and harness-controlled inference experiments use Claude Sonnet-4.6 as the fixed base model.
We choose Sonnet-4.6 because it is a strong frontier model with broad capabilities in reasoning, coding, long-context understanding, computer use, and agentic workflows~\citep{anthropic_sonnet46_page,anthropic_sonnet46_system_card}.
Importantly, the model is kept fixed throughout our experiments: we do not update its parameters, fine-tune it on benchmark-specific data, or replace it with a stronger model at test time.
Therefore, any improvement from pass@1 to the metacognitive harness reflects a change in the inference-time control procedure rather than a change in the underlying model.

\paragraph{Leaderboard reference models.}
For context, we compare our results against representative top leaderboard models reported for each benchmark snapshot.
These models are used only as external reference points; they are not used by our harness.
The OpenAI models appearing in our tables include GPT-5.2, GPT-5-mini, o3, and o4-mini~\citep{openai_all_models,openai_gpt52,openai_gpt5mini,openai_o3,openai_o4mini}.
GPT-5.2 and GPT-5-mini are GPT-5-family models, while o3 and o4-mini are reasoning-oriented o-series models; suffixes such as ``H'' and ``M'' in our tables denote the high- and medium-effort settings reported by the corresponding leaderboard or model interface, rather than separate model families.
We also include Claude Opus-4.6 as an Anthropic reference model~\citep{anthropic_opus46_system_card,anthropic_system_cards}.
Opus-4.6 is an Opus-class Claude model designed for high-capability reasoning, coding, agentic search, and knowledge-work tasks.
From Google, we include Gemini-3-Pro and Gemini-2.5-Pro~\citep{google_gemini3_guide,google_gemini25pro,google_gemini_models}.
Gemini-3-Pro is used as a leaderboard-snapshot reference model for advanced multimodal and long-context reasoning, while Gemini-2.5-Pro is a strong thinking model for code, math, STEM, long-context analysis, and document-level reasoning.
Because model availability and naming conventions can change over time, these leaderboard entries should be interpreted as benchmark-snapshot comparisons rather than claims about the current availability of any particular API model.

\paragraph{Open-weight multimodal reference model.}
For open-weight multimodal comparison, we include Qwen2.5-VL-72B-Instruct~\citep{qwen25vl_72b_hf,qwen25vl_blog,qwen25vl_techreport}.
Qwen2.5-VL-72B-Instruct is the instruction-tuned 72B model in the Qwen2.5-VL family.
The Qwen2.5-VL series is designed for image and video understanding, document and diagram parsing, structured visual grounding, object localization, and multimodal reasoning.
In our tables, this model appears only as a leaderboard reference point for R-Bench-V; it is not used as the base model for our harness experiments.

\paragraph{Interpretation of model comparisons.}
Our goal is not to claim that Sonnet-4.6 is intrinsically stronger than every reference model under all possible settings.
Instead, the comparison is intended to evaluate whether a fixed strong model can better exploit its own reasoning capability when equipped with an explicit metacognitive control harness.
This distinction is important: standard pass@1 inference commits to a single trajectory, whereas our harness uses the model's own pre-solve and post-solve self-evaluation signals to decide when to retry, which attempts to trust, and how to aggregate candidate answers.
Thus, the leaderboard comparisons serve as context for the absolute strength of the resulting system, while the pass@1-to-harness comparison isolates the effect of metacognitive control on the same base model.

\section{Experiment Setting Details}
\label{app:exp_details}

\paragraph{Metacognitive diagnosis set.}
For metacognitive diagnosis, we use a fixed 100-example anchor set spanning text, code, and multimodal reasoning.
This set is disjoint from the downstream benchmark evaluation sets and is used only to assess whether a model exposes self-evaluation signals that are reliable enough to support control.
For each example, we collect a pre-solve FOK score, a post-solve JOL score, and the correctness label of the corresponding attempt.

\paragraph{Diagnosis metrics and verdicts.}
We evaluate each model along two axes.
Discrimination measures whether higher-confidence examples are more likely to be correct, using AUROC.
Calibration measures whether confidence magnitudes match empirical correctness, using ECE.
We report these metrics for FOK, JOL, and a joint signal obtained from a lightweight per-model classifier over $(\mathrm{FOK}, \mathrm{JOL})$.
These metrics determine whether a model is suitable for metacognitive harnessing.

\paragraph{Controller fitting.}
The harness controller maps the metacognitive state $(\mathrm{FOK}, \mathrm{JOL})$ to a retry/stop decision.
In the main system, this controller is implemented as a per-model SVM trained on the diagnosis set.
The goal is not to impose a universal threshold, but to learn a model-specific decision boundary that converts raw self-assessment scores into a calibrated control signal.
We compare this learned controller against simpler alternatives, including hand-crafted confidence rules and signal-drop variants such as FOK-only and JOL-only control.

\paragraph{Reasoning-time control and context management.}
At test time, the harness queries one pre-solve FOK score and a post-solve JOL score after each attempt.
The controller then decides whether to trust the current attempt or allocate another retry.
When retrying, the next attempt receives only a compact metacognitive history, consisting of prior answers, JOL scores, and short JOL reasons.
Previous full reasoning traces are intentionally excluded to avoid anchoring later attempts to earlier failed trajectories.

\paragraph{Final aggregation.}
When multiple attempts are generated, final aggregation is performed separately from metacognitive control.
The aggregator receives the reasoning traces and answers from all attempts, but not FOK, JOL, or JOL reasons.
This design reflects our empirical finding that confidence is useful for deciding whether more computation is needed across problems, but is much less informative for distinguishing among attempts within the same problem.

\paragraph{Ablation protocol.}
Controlled ablations are conducted on a separate 100-example subset randomly sampled from the testing data, using Sonnet-4.6 as the fixed base model.
These ablations isolate the contribution of individual design choices, including FOK and JOL removal, replacement of the learned controller with simpler rules, random retry under matched budgets, simplified aggregation strategies, and alternative context-management variants.
The main benchmark results are always reported separately on the full evaluation splits.

\paragraph{Budget accounting.}
For baseline comparisons, we match or closely control the amount of inference-time computation whenever possible.
Vertical scaling baselines spend additional computation within a single trajectory, while parallel scaling baselines spend computation by generating multiple independent attempts.
The metacognitive harness differs in that it allocates this computation adaptively through retry and stopping decisions conditioned on the model's self-monitoring signals.

\paragraph{Oracle analysis.}
When a method generates multiple attempts, we report oracle@$K$ to measure the reachable capacity of its exploration policy.
Oracle@$K$ counts an example as correct if any of the generated attempts is correct.
This metric is used only as an upper bound on the quality of the explored candidate set, not as a practical inference procedure.

\section{Detailed Per-Benchmark Results}
\label{app:per_benchmark_results}

The main paper reports pooled accuracy over all evaluation examples to provide a single summary of test-time scaling effectiveness.
Here we provide the corresponding per-benchmark results.
For each benchmark, we report accuracy, absolute gain over Sonnet-4.6 pass@1, oracle accuracy, average number of attempts, and average cost per question.
Oracle accuracy measures whether at least one generated attempt among the candidate set is correct.
Avg. $K$ denotes the realized average number of solution attempts per question; for adaptive methods, this value can be smaller than the maximum allowed budget due to early stopping.
Cost is computed as the total inference cost divided by the number of evaluated questions.

\paragraph{HLE-Verified.}
Table~\ref{tab:app_hle_results} reports detailed results on HLE-Verified.
The metacognitive harness improves Sonnet-4.6 from 48.0 to 60.0, yielding a +12.0 point gain.
Compared with vertical scaling methods such as Self-Refine and budget forcing, the harness provides a larger improvement because it does not simply extend a single trajectory.
Compared with parallel scaling methods such as verifier or aggregator selection, the harness uses metacognitive signals to allocate retries and select among attempts more selectively.

\begin{table}[h]
\centering
\small
\setlength{\tabcolsep}{5.0pt}
\renewcommand{\arraystretch}{1.12}
\caption{Detailed results on HLE-Verified.}
\label{tab:app_hle_results}
\begin{tabular*}{\linewidth}{@{\extracolsep{\fill}}lccccc@{}}
\toprule
Method
& Acc.
& Gain
& Oracle Acc.
& Avg. $K$
& Avg. Cost / Q. \\
\midrule
Sonnet-4.6 Pass@1
& 48.0
& --
& --
& 1.0
& \$0.50 \\

Self-Refine
& 53.0
& +5.0
& 57.7
& 1.7
& \$0.41 \\

Budget Forcing
& 51.0
& +3.0
& 51.0
& 1.0
& \$0.60 \\

Verifier Reranking
& 52.0
& +4.0
& 63.0
& 4.0
& \$2.21\\

Aggregator
& 57.0
& +9.0
& 63.0
& 4.0
& \$2.31 \\

\rowcolor{MCOursGreen}
Metacognitive Harness
& \textbf{60.0}
& \textbf{+12.0}
& \textbf{64.0}
& \textbf{1.4}
& \textbf{\$0.98} \\
\bottomrule
\end{tabular*}
\end{table}

\paragraph{LiveCodeBench v6.}
Table~\ref{tab:app_lcb_results} reports detailed results on LiveCodeBench v6.
The harness improves Sonnet-4.6 from 74.3 to 84.3 overall, yielding a +10.0 point gain.
This improvement is especially meaningful because LiveCodeBench requires both high-level algorithmic reasoning and executable implementation.
The harness outperforms both vertical scaling baselines and parallel selection baselines, suggesting that metacognitive control provides a more targeted use of test-time computation than simply extending or sampling reasoning trajectories.

\begin{table}[h]
\centering
\small
\setlength{\tabcolsep}{5.0pt}
\renewcommand{\arraystretch}{1.12}
\caption{Detailed results on LiveCodeBench v6.}
\label{tab:app_lcb_results}
\begin{tabular*}{\linewidth}{@{\extracolsep{\fill}}lccccc@{}}
\toprule
Method
& Acc.
& Gain
& Oracle Acc.
& Avg. $K$
& Avg. Cost / Q. \\
\midrule
Sonnet-4.6 Pass@1
& 74.3
& --
& --
& 1.0
& \$0.24 \\

Self-Refine
& 80.6
& +6.3
& 85.5
& 1.1
& \$0.32 \\

Budget Forcing
& 80.0
& +5.7
& 80.0
& 1.0
& \$0.50 \\

Verifier Reranking
& 78.9
& +4.6
& 83.4
& 4.0
& \$0.84 \\

Aggregator
& 83.4
& +9.1
& 83.4
& 4.0
& \$0.94 \\

\rowcolor{MCOursGreen}
Metacognitive Harness
& \textbf{84.3}
& \textbf{+10.0}
& \textbf{86.0}
& \textbf{2.6}
& \textbf{\$0.56} \\
\bottomrule
\end{tabular*}
\end{table}

\paragraph{R-Bench-V.}
Table~\ref{tab:app_rbv_results} reports detailed results on R-Bench-V.
The harness improves Sonnet-4.6 from 36.1 to 41.1 overall, yielding a +5.0 point gain.
Unlike HLE and LiveCodeBench, R-Bench-V requires visual grounding in addition to reasoning.
The consistent improvement on this benchmark suggests that metacognitive control is not limited to text-only reasoning, but can also improve multimodal reasoning when the model's self-evaluation signals are sufficiently informative.

\begin{table}[h]
\centering
\small
\setlength{\tabcolsep}{5.0pt}
\renewcommand{\arraystretch}{1.12}
\caption{Detailed results on R-Bench-V.}
\label{tab:app_rbv_results}
\begin{tabular*}{\linewidth}{@{\extracolsep{\fill}}lccccc@{}}
\toprule
Method
& Acc.
& Gain
& Oracle Acc.
& Avg. $K$
& Avg. Cost / Q. \\
\midrule
Sonnet-4.6 Pass@1
& 36.1
& --
& --
& 1.0
& \$0.15 \\

Self-Refine
& 38.2
& +2.1
& 41.7
& 1.6
& \$0.55 \\

Budget Forcing
& 38.2
& +2.1
& 38.2
& 1.0
& \$0.80 \\

Verifier Reranking
& 34.0
& -2.1
& 47.3
& 4.0
& \$0.80\\

Aggregator
& 36.5
& +0.4
& 47.3
& 4.0
& \$0.90\\

\rowcolor{MCOursGreen}
Metacognitive Harness
& \textbf{41.1}
& \textbf{+5.0}
& \textbf{49.2}
& \textbf{3.2}
& \textbf{\$0.72} \\
\bottomrule
\end{tabular*}
\end{table}

%

\section{Case Studies: Three Regimes of the Metacognitive Harness}
\label{sec:case_studies}

We walk through three short trajectories produced by the deployed harness
(Sonnet-4.6 solver, joint-SVM controller on $(1{-}\mathrm{FOK},1{-}\mathrm{JOL})$,
hybrid aggregator). The three cases are picked to illustrate the
\emph{different} ways the controller can spend (or save) its retry budget:
exhausting all four samples and relying on the aggregator (Sec.~\ref{sec:case_hle}),
stopping at $K{=}1$ because the post-solve confidence is decisive
(Sec.~\ref{sec:case_lcb}), and a single corrective retry on a visual problem
(Sec.~\ref{sec:case_rbenchv}). For each case we report the raw FOK and JOL
scores, the per-attempt retry signal $(1{-}\mathrm{JOL})\!\cdot\!\mathrm{FOK}$
used as an interpretable proxy for the SVM's stop probability, and excerpts
of the actual model reasoning across attempts.

\subsection{Case 1 (HLE, knowledge retrieval): low confidence, full budget,
aggregation recovers truth}
\label{sec:case_hle}

\paragraph{Question (HLE, id \texttt{6725145d\dots94d6}).}
\textit{``What is the current Vietnamese province where Ming general Mu Sheng
experienced his first major defeat?''}
\quad \textbf{Gold:} Nam Dinh.

\paragraph{Metacognitive scores.}
$\mathrm{FOK}{=}0.30$,\; $\mathrm{JOL}{=}0.18$.\;
Both signals are pessimistic: the model judges that it is unlikely to know
the answer \emph{before} solving (FOK), and unlikely to have produced a
correct answer \emph{after} the first attempt (JOL). The retry signal is
$(1{-}0.18)\cdot 0.30 \approx 0.25$ on the hand-tuned scale, but the
calibrated SVM places three of the four post-attempt states above its stop
threshold; the harness therefore consumes the full $K{=}4$ budget. The raw
sequence of retry-signal-equivalent values is $1.31 \to 1.18 \to 1.27$ (all
above $\tau$).

\paragraph{Per-attempt reasoning.}
The four samples drift across plausible Vietnamese provinces because the
model anchors on a different period of the Ming-Vietnam wars each time.
\begin{enumerate}
  \item \textit{Attempt 1: Nghe An (incorrect).} The solver assumes the
        defeat occurred during the late Lam Son uprising (c.~1424--1425) and
        guesses the province where those campaigns were concentrated.
  \item \textit{Attempt 2: Nam Dinh (correct).} The solver re-anchors on the
        \emph{earlier} phase of the occupation, identifies the Battle of Bo
        Co (1408) as Mu Sheng's first major defeat, and correctly locates
        the engagement in the coastal Red-River-delta area corresponding to
        modern Nam Dinh.
  \item \textit{Attempt 3: Ninh Binh (incorrect).} Now the solver fixes on
        the Battle of Bo Co but reasons \emph{geographically} from the
        Day River estuary, which it places in Ninh Binh.
  \item \textit{Attempt 4: Thai Binh (incorrect).} The solver explicitly
        notes that the previous three attempts disagreed and tries yet
        another distributary of the Red River delta.
\end{enumerate}

\paragraph{Aggregation and outcome.}
Because the four candidates are split four ways, string-consensus does not
fire and the question is not a code task, so the hybrid aggregator falls
through to the select-attempt stage. Given the four candidate
answers and their reasoning traces, the aggregator selects Attempt 2
(\textbf{Nam Dinh}), which matches the gold answer.

\paragraph{Takeaway.}
This is the regime the harness is designed for: the model's own metacognition
correctly reports that it does not know, the controller refuses to stop, and
the aggregator recovers truth from one good sample buried among three
distractors. A confidence-blind sampler that picks the last attempt would
have reported Thai Binh; a self-consistency vote would have ended in a tie.

\subsection{Case 2 (LiveCodeBench, code generation): low FOK overridden by
high JOL, harness stops at $K{=}1$}
\label{sec:case_lcb}

\paragraph{Question (LCB, AtCoder \texttt{abc370\_a}, easy).}
Read two binary inputs $L,R\in\{0,1\}$ representing whether the left and
right hands are raised; print \textsc{Yes} if only the left hand is up,
\textsc{No} if only the right is up, and \textsc{Invalid} otherwise.
\textbf{Gold:} a three-branch conditional that passes all public tests.

\paragraph{Metacognitive scores.}
$\mathrm{FOK}{=}0.30$,\; $\mathrm{JOL}{=}0.99$.\;
The pre-solve FOK is low because the model has no useful prior on the
specific AtCoder ID; FOK is calibrated against ``do I know this problem'',
not against ``can I solve a problem of this kind''. After the first attempt,
however, JOL is essentially saturated: the solver has written a small
\verb|if/elif/else| program that exhaustively covers the four input cases
and matches both worked examples. The retry signal therefore collapses to
$(1{-}0.99)\cdot 0.30 \approx 0.003$, well below~$\tau$, and the harness
stops at $K{=}1$.

\paragraph{Per-attempt reasoning (Attempt 1, only).}
The solver enumerates the four input cases and emits the canonical solution:
\begin{verbatim}
L, R = map(int, input().split())
if L == 1 and R == 0: print("Yes")
elif L == 0 and R == 1: print("No")
else: print("Invalid")
\end{verbatim}
The reasoning trace explicitly cross-checks each branch against the two
sample inputs in the problem statement before committing.

\paragraph{Outcome.}
Single-attempt answer is correct; total cost is one solver call.

\paragraph{Takeaway.}
This case justifies keeping JOL in the controller even though its marginal
1D AUROC is similar to the joint signal: pre-solve FOK alone would have kept
retrying on a problem the model had already solved, paying $4\times$ the
compute for no accuracy gain. The early-stop hit-rate column of
Table~\ref{tab:ablation_harness} (68.9\%) is dominated by cases of this
shape.

\subsection{Case 3 (R-Bench-V Physics, visual reasoning): one retry corrects
a topology error}
\label{sec:case_rbenchv}

\paragraph{Question (R-Bench-V, \texttt{physics\_149}).}
\textit{``Twenty identical resistors, each with resistance $R$, are connected
as shown in the diagram; determine the equivalent resistance between points
$A$ and $B$.''}
\quad \textbf{Gold:} $R_{\mathrm{eq}}{=}2R$.

\paragraph{Metacognitive scores.}
$\mathrm{FOK}{=}0.45$,\; $\mathrm{JOL}{=}0.72$.\; Both signals are
intermediate. The solver is moderately confident the topology is solvable
by symmetry, but not confident the topology was \emph{read} correctly from
the figure. The retry signal sequence is $0.85 \to 0.55$: above threshold
after attempt~1 (retry), below threshold after attempt~2 (stop). The
controller therefore samples exactly twice.

\paragraph{Per-attempt reasoning.}
\begin{enumerate}
  \item \textit{Attempt 1: $R_{\mathrm{eq}}{=}4R$ (incorrect).} The solver
        misreads the diagram as a $4{\times}2$ grid of diamond cells and
        applies a series-of-bridges decomposition. The post-solve JOL
        ($0.72$) is \emph{below} the high-confidence regime, signalling
        that the topology assumption is shaky.
  \item \textit{Attempt 2: $R_{\mathrm{eq}}{=}2R$ (correct).} The solver
        re-counts and now reads a $3{\times}2$ diamond lattice, exploits
        top--bottom symmetry to merge equipotential nodes, and computes
        $R_{\mathrm{eq}}{=}2R$. Post-solve JOL rises sharply, the retry
        signal drops below $\tau$, and the harness stops.
\end{enumerate}

\paragraph{Outcome.}
Final answer $2R$, correct, two attempts.

\paragraph{Takeaway.}
This is the directed-retry regime: the controller spends exactly one extra
attempt on a problem where the first sample's reasoning was internally
consistent but rested on a wrong premise (the lattice shape). The next
attempt is given a fresh budget to re-examine the figure and the now-high
JOL terminates the trajectory at the right moment. A fixed-$K{=}4$ baseline
would have produced the same final answer at twice the cost; a fixed-$K{=}1$
baseline would have reported $4R$.

\subsection{Summary across the three cases}

The three trajectories cover the three operating points of the controller:
(i) high-uncertainty problems where the harness should and does spend its
full budget and lean on the aggregator (Case~1);
(ii) easy or already-solved problems where the post-solve JOL kills the
budget (Case~2);
(iii) intermediate problems where one retry, directed by an elevated retry
signal, is exactly the right amount of extra compute (Case~3).
The common thread is that the stop decision in each case is driven by a
\emph{combination} of FOK and JOL --- in Case~2 the post-solve JOL is the
load-bearing signal, in Case~3 the gap between FOK and JOL is the
load-bearing signal, and in Case~1 both signals jointly veto stopping.
This matches the ablation in Table~\ref{tab:ablation_harness}: dropping
either signal hurts the average budget $\bar K$, and dropping the SVM in
favour of the hand-tuned $(1{-}\mathrm{JOL})\cdot\mathrm{FOK}$ rule hurts
the early-stop hit rate.

\section{Limitations}

The current harness is not model-universal: its effectiveness depends on the base model exposing metacognitive signals that are sufficiently discriminative and calibratable.
Our main end-to-end evaluation therefore focuses on Sonnet-4.6, which exhibits the strongest metacognitive diagnosis in our model panel.
Other promising models, such as DeepSeek-V4-Pro and Gemini-3-Flash, may also benefit from harnessing, but a broader model sweep is left to future work.
In addition, the controller is fitted on a small anchor set, which keeps diagnosis lightweight but may limit robustness under distribution shift.
Finally, our evaluation focuses on benchmark reasoning tasks rather than long-horizon agent settings, so the effectiveness of metacognitive control in richer agent workflows remains to be tested.

\end{document}